\documentclass[final]{cvpr}

\usepackage{times}
\usepackage{epsfig}
\usepackage{graphicx}
\usepackage{amsmath}
\usepackage{amssymb}

\DeclareMathOperator*{\argmax}{arg\,max}
\graphicspath{{images/}{../images/}}
\usepackage{subfig}
\usepackage{algorithm, algpseudocode}
\usepackage{subfiles}
\usepackage{chngcntr}
\usepackage[normalem]{ulem}

\def\cvprPaperID{3996} % *** Enter the CVPR Paper ID here
\def\confYear{CVPR 2021}
\begin{document}

%%%%%%%%% TITLE
\title{Sequential Graph Convolutional Network for Active Learning}

\author{Razvan Caramalau$^1$, Binod Bhattarai$^1$ and Tae-Kyun Kim$^{1,2}$ \\
$^1$Imperial College London, UK\\
$^2$KAIST, South Korea\\
{\tt\small \{ r.caramalau18, b.bhattarai, tk.kim\}@imperial.ac.uk}
% For a paper whose authors are all at the same institution,
% omit the following lines up until the closing ``}''.
% Additional authors and addresses can be added with ``\and'',
% just like the second author.
% To save space, use either the email address or home page, not both
% \and
% Tae-Kyun Kim \\
% KAIST, South Korea\\
}

\maketitle

%%%%%%%%% ABSTRACT
\begin{abstract}
We propose a novel pool-based Active Learning frame-work constructed on a sequential Graph Convolution Net-work (GCN). Each image’s feature from a pool of data rep-resents a node in the graph and the edges encode their similarities.   With a  small number of randomly sampled images as seed labelled examples,  we learn the parameters of the graph to distinguish labelled vs unlabelled nodes by minimising the binary cross-entropy loss.   GCN performs message-passing operations between the nodes, and hence, induces similar representations of the strongly associated nodes. We exploit these characteristics of GCN to select the unlabelled examples which are sufficiently different from la-belled ones.  To this end, we utilise the graph node embed-dings and their confidence scores and adapt sampling techniques such as CoreSet and uncertainty-based methods to query the nodes.  We flip the label of newly queried nodes from unlabelled to labelled, re-train the learner to optimise the downstream task and the graph to minimise its modified objective.  We continue this process within a fixed budget.   We evaluate our method on  6  different benchmarks: 4 real image classification, 1 depth-based hand pose estimation and 1 synthetic RGB image classification datasets. Our method outperforms several competitive baselines such as VAAL, Learning Loss, CoreSet and attains the new state-of-the-art performance on multiple applications.

\end{abstract}
%%%%%%%%% BODY TEXT
% \section{Introduction}
\section{Introduction}
\label{sec:intro}

Deep learning has shown great advancements in several computer vision tasks such as 
image classification~\cite{he2016deep,krizhevsky2012imagenet} and 3D Hand Pose Estimation (HPE) ~\cite{deepprior, qiye, deepprior++}.
This has been possible due to the availability of both the powerful computing infrastructure and the
large-scale  datasets. 
% Some of the large-scale benchmarks 
% such as CIFAR-10, CIFAR-100~\cite{cifar}, NYU Hand~\cite{nyu}, and BigHand2.2M\cite{bighand}  are in the orders of $10^5$.
% with the cost of large-scale annotated data sets. 
Data annotation is a time-consuming task, needs experts and is also expensive.
This gets even more challenging to some of the specific domains such as robotics or medical image analysis. Moreover, while optimizing deep neural network architectures, a gap is present concerning the representativeness of the data~\cite{data_redundant}. To overcome these issues, \emph{active learning}~\cite{Gal2017DeepDatab,near2} has been successfully deployed to efficiently select the most meaningful samples. 
\begin{figure}
    \centering
    \includegraphics[trim=0cm 0cm 0cm 0.0cm, clip, width=0.47\textwidth]{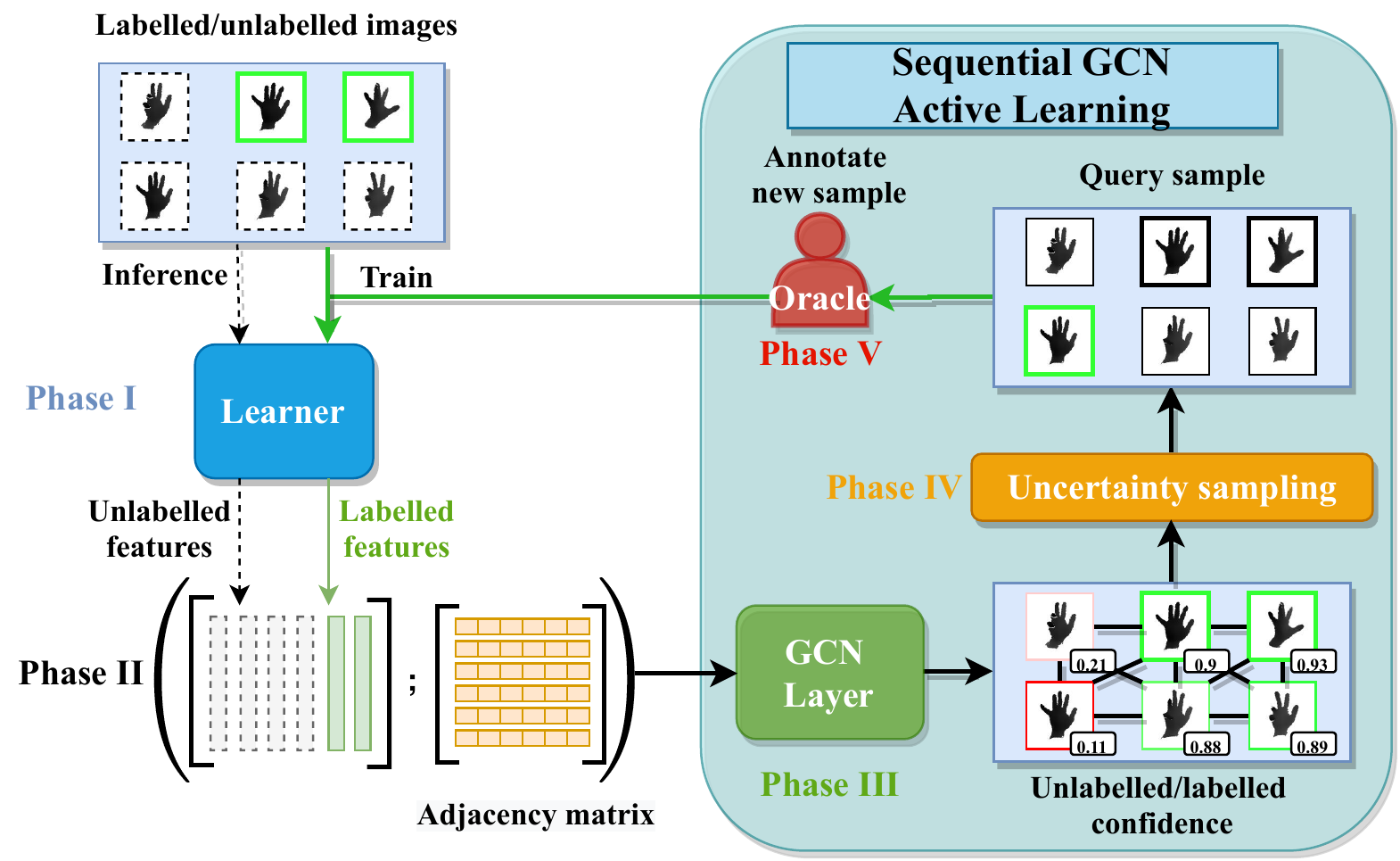}
    \caption{The diagram outlines the proposed pipeline. \textbf{Phase I}: Train the learner to minimise the objective of downstream task from the available annotations, \textbf{Phase II}: Construct a graph with the representations of images extracted from the learner as vertices and their similarities as edges, \textbf{Phase III}: Apply GCN layers to minimise binary cross-entropy between labelled and unlabelled, \textbf{Phase IV}: Apply uncertainty sampling to select unlabelled examples to query their labels, \textbf{Phase V}: Annotate the selection, populate the number of labelled examples and repeat the cycle.}
    \label{fig:prposed_pipeline}    
\end{figure}

Essentially, there are three distinct components in any Active Learning (AL) framework. 
These components are \emph{learner, sampler}, and \emph{annotator}. In brief, a \emph{learner} is a model 
trained to minimize the objective of the target task. The \emph{sampler} is designed to select 
the representative unlabelled examples within a fixed budget to deliver the highest performance.
Finally, an \emph{annotator} labels the queried data for learner re-training. Based on
the relationship between learner and sampler, AL frameworks can be categorised into 
two major groups: task-dependent and task-agnostic. Task-dependents are the ones where 
the sampler is specially designed for a specific task of the learner.
Majority of the previous works in active learning~\cite{cortes1995support, Gal2017DeepDatab,AbelRegionalLearning,Wu2019ActivePropagation,Cai2017ActiveEmbedding, anbr} are task-dependent in nature.
% implant both the selection method and the learner, the two major components of the active learning framework, together. 
In other words, the sampling function is dependent on the objective of the learner.  
This design limits the model to become optimal for a specific type of task while suffering from its scalability problem. 
Some of the recent works such as
VAAL~\cite{Sinha2019VariationalLearning} and Learning
Loss~\cite{Yoo2019LearningLearning} tackle such a problem.
VAAL trains a variational auto-encoder (VAE) that learns a latent space for better discrimination 
between labelled and unlabelled images in an adversarial manner.  
% This ignores such relationship between the images. 
% \sout{Similarly, Learning Loss introduces extra loss 
% function in a self-supervised fashion and ignores the aspect of the data relationship.}
Similarly, Learning Loss introduces a separate loss-prediction module to
be trained together with the learner.
% so you think learning loss is not a self-supervised method? I consider that a self-supervised way of training even they have not mentioned in the paper.}
% \textcolor{red}{The major drawback of these approaches is not being
% able to exploit the relative relation between within the labelled, unlabelled
% and across the labelled and unlabelled data. And, also there is no communicating mechanism between 
% the labelled, unlabelled and candidate examples for querying their labels. Not clear}
The major drawback of these approaches is the lack of a mechanism that exploits the 
correlation between the labelled and the unlabelled images. Moreover, VAAL has 
no way to communicate between the learner and the sampler.  
% no passing communication with the learner by making it disconnected from the downstream understanding.
Graph Convolutional Networks(GCNs)~\cite{kipf2016semi,bronstein2017geometric} are capable 
of sharing information between the nodes via message-passing operations.
In the AL domain, the application of GCNs \cite{Cai2017ActiveEmbedding, anbr, AbelRegionalLearning, Wu2019ActivePropagation} 
is also slowly getting priority. However, these methods are applied only to 
specific kind of datasets i.e. graph-structured data such as Cora, CiteSeer, 
and PubMed~\cite{graph_datasets}. In this work, we are exploring
the image domain beyond graph-structured data.

\begin{figure*}
    \centering
    \includegraphics[trim=0.0cm 0.0cm 0.0cm 0.0cm, clip, width=0.9\textwidth]{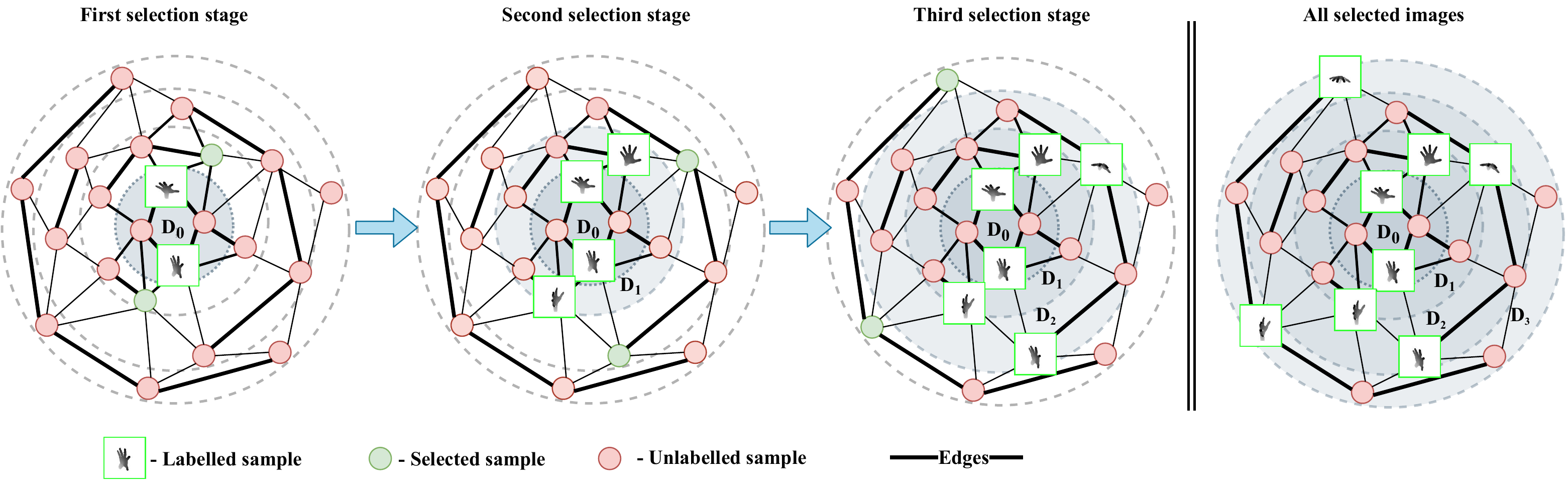}
    \caption{ This Figure simulates the sampling behaviour of the proposed \textbf{UncertainGCN} 
    sampler at its first three selection stages. We run a toy experiment just taking 100 examples
    from ICVL~\cite{icvl} Hand Pose dataset. Each node is initialised by the features extracted 
    from the learner and edges
    capture their relation. Each concentric circle represents a cluster of strongly connected nodes.
    Here, in our case, a group of images having similar viewpoints are in a concentric circle.
    Considering two labelled examples as seed labelled examples in the centre-most circle, in the first selection stage, our algorithm selects samples from another concentric circle which is out-of-distribution than selecting the remaining examples from the innermost circle. Similarly, 
    in the second stage, our sampler chooses images residing in another outer concentric circle which are sufficiently different from those selected before.}
    \label{fig:sampling_simulation}    

\end{figure*}

To address the above-mentioned issues, we propose a sequential GCN for Active Learning in a task-agnostic manner. 
% In comparison to previous works, we propose to train the framework in a 
% task-agnostic manner. 
Figure~\ref{fig:prposed_pipeline} shows the pipeline of the proposed method. 
In the Figure, Phase I implements the learner. This is a model trained to minimize the 
objective of the downstream task. Phase II, III and IV compose our sampler where 
we deploy the GCN and apply the sampling techniques on
graph-induced node embeddings and their confidence scores. Finally, in Phase V, the 
selected unlabelled examples are sent for annotation.
At the initial stage, the learner is trained with a small number of seed labelled examples.
We extract the features of both labelled and unlabelled images from the learner parameters.
During Phase II, we construct a graph where features are used to initialise the nodes of the graph 
and similarities represent the edges. Unlike VAAL\cite{Sinha2019VariationalLearning}, the initialisation of the nodes
by the features extracted from the learner creates an opportunity to inherit 
uncertainties to the sampler. This graph is passed through GCN layers (Phase III) 
and the parameters of the graph are learned to
identify the nodes of labelled vs unlabelled example. This objective of the sampler is
independent of the one from the learner. % different objectives don't necessarily result in task-agnostic schemes
We convolve on the graph which does message-passing operations between the nodes to 
induce the higher-order representations. The graph embedding of any image depends primarily 
upon the initial representation and the associated neighbourhood nodes. Thus, the images
bearing similar semantic and neighbourhood
structure end up inducing close representations which will play a key 
role in identifying the sufficiently different 
unlabelled examples from the labelled ones. The nodes after convolutions are classified 
as labelled or unlabelled. We sort the examples based on the confidence score, apply an uncertainty sampling
approach (Phase IV), and send the selected examples to query their labels(Phase V). We called this sampling 
method as \textbf{UncertainGCN}. Figure~\ref{fig:sampling_simulation} simulates the 
UncertainGCN sampling technique.
Furthermore, we adapt the higher-order graph node information under the CoreSet \cite{Sener2017ActiveApproach} 
for a new sampling technique by introducing latent space distancing. In principle, CoreSet ~\cite{Sener2017ActiveApproach} uses risk minimisation between core-sets on the learner feature space while we
employ this operation over GCN features. We called this sampling technique as \textbf{CoreGCN}.
Our method has a clear advantage due to the aforementioned strengths of the GCN which is demonstrated by both 
the quantitative and qualitative experiments (see Section \ref{experiments}).
Traversing from Phase I to Phase V as shown in Figure~\ref{fig:prposed_pipeline} 
completes a cycle. In the next iteration, we flip the label of annotated 
examples from unlabelled to labelled and re-train the whole framework.

We evaluated our sampler on four challenging real domain image classification benchmarks, one depth-based dataset for 3D HPE and a synthetic
image classification benchmark. We compared with several competitive sampling baselines and existing state-of-the-arts methods including CoreSet, VAAL and Learning Loss. From both the quantitative and the qualitative comparisons, our proposed framework is more accurate than existing methods.

\section{Related Works}
\label{related_works}
\noindent \textbf{Model-based methods.} Recently, a new category of methods is explored in the active learning paradigm where a separate model from the learner is trained for selecting a sub-set of the most representative data. Our method is based on this category. One of the first approaches \cite{Yoo2019LearningLearning} attached a loss-learning module so that loss can be predicted offline for the unlabelled samples. In \cite{Sinha2019VariationalLearning}, another task-agnostic solution deploys a variational auto-encoder (VAE) to map the available data on a latent space. Thus, a discriminator is trained in an adversarial manner to classify labelled from unlabelled. The advantage of our method over this approach is the exploitation of the relative relationship between the 
 examples by sharing information through message-passing operations in GCN.

\noindent \textbf{GCNs in active learning.} 
% New possibilities in this domain were opened by introducing the GCNs \cite{kipf2016semi}. A recent survey \cite{Wu2019ANetworks} gathered the advancements for this architecture.
GCNs ~\cite{kipf2016semi} have opened new active learning methods that have been successfully applied in~\cite{AbelRegionalLearning,Wu2019ActivePropagation,Cai2017ActiveEmbedding, anbr}. In comparison to these methods, our approach has distinguished learner and sampler. It makes our approach task-agnostic and also
gets benefited from model-based methods mentioned just before. Moreover, none of these methods is trained in a sequential manner. 
% However, none of the existing methods been applied to computer vision problems.
% However, all of them integrated the selection mechanism on the GCN learner. Thus, their active learning framework remains dependent on the graph architecture.
~\cite{Wu2019ActivePropagation} proposes K-Medoids clustering for the feature propagation between the labelled and unlabelled nodes. A regional uncertainty algorithm is presented in \cite{AbelRegionalLearning} by extending the PageRank \cite{ilprints422} algorithm to the active learning problem. Similarly, \cite{Cai2017ActiveEmbedding} combines node uncertainty with graph centrality for selecting the new samples. A more complex algorithm is introduced in \cite{anbr} where a reinforcement scheme with multi-armed bandits decides between the three query measurements from \cite{Cai2017ActiveEmbedding}. 
% Similarly, \cite{garcia2017fewshot} applies GCN few-shot active learning while our method is designed for supervised learning algorithms.
% They draw a short active learning experiment against random sampling. Our proposed methods differentiate from their sampling methodology that integrates a softmax attention layer on the GCN features. Moreover, they lack of comprehensive experiments and sequential processes. 
However, these works~\cite{Cai2017ActiveEmbedding, anbr, Wu2019ActivePropagation} derive their selection mechanism on the assumption of a Graph learner. This does not make them directly comparable with our proposal where a GCN is trained separately for a different objective function than the learner.

% \textcolor{blue}{In this Section, we cover extensive sets literatures on active learning. Broadly, the existing methods can be categorised into  ... groups. Our method falls under, ... group. Hence, we discuss works on .. category
% in more details.}

%\paragraph{Active Learning.} This branch of machine learning have shown remarkably developments over the past two decades. Starting from classical methods suitable for linear regression, SVM \cite{cortes1995support} (Support Vector Machines), committee machines \cite{comm}, active learning has gained popularity recently for deep neural networks. The taxonomy of the first principles has been gathered in \cite{settles.tr09}, representing the stepping stone for the research in this field. Nowadays, as training deep learning architectures is time-demanding, pool-based scenarios presented in \cite{settles.tr09} are commonly applied.
\noindent \textbf{Uncertainty-based methods.} Earlier techniques for sampling unlabelled data have been explored through uncertainty exploration of the convolutional neural networks (CNNs). A Bayesian approximation introduced in \cite{Gal2016DropoutGhahramani} produce meaningful uncertainty measurements by variational inference of a Monte Carlo Dropout (MC Dropout) adapted architecture. Hence, it is successfully integrated into active learning by \cite{Gal2017DeepDatab,Houlsby2011BayesianLearning,Kirsch2019BatchBALD:Learning, Pinsler2019BayesianApproximation}. 
With the rise of GPU computation power, \cite{BeluchBcai2018TheClassification} ensemble-based method outperformed MC Dropout.%Despite this, querying in this manner might not be advisable for some applications.
\\
\textbf{Geometric-based methods.} Although there have been studies exploring the data space through the representations of the learning model (\cite{near2,coresvm,Har-Peled2007}), the first work applying it for CNNs as an active learning problem, CoreSet, has been presented in \cite{Sener2017ActiveApproach}. The key principle depends on minimising the difference between the loss of a labelled set and a small unlabelled subset through a geometric-defined bound. We furthermore represent this baseline in our experiments as it successfully over-passed the uncertainty-based ones. 
% core-set k-means. k-medoids, exploring model's unlabeled feature space

%if following works are not published, no need to cite them
% \st{Moving into this direction, a novel approach \cite{Kim2020Task-AwareLearning} has combined this adversarial training with the learning-loss module by ranking the generated latent features according to the task learner} 
% Furthermore, following the same principle, a state re-labelling mechanism has been proposed in \cite{Zhang2020State-RelabelingLearning} where an online uncertainty module refines the decision of the discriminator. 

% \section{Methodology}
\section{Method}
% \begin{figure*}
%     \centering
%     \includegraphics[trim=0cm 11cm 9cm 0cm, clip, width=0.85\linewidth]{images/diagram.pdf}
%     \caption{Schematic diagram showing the proposed pipeline. Here, \emph{Image classifier} depicts our learner and \emph{GCN} block represents the selection framework which are the key components of active learning pipeline. For the sanity, we are not showing the weakly edges.}
%     \label{fig:pipeline}
% \label{pipeline}
\def\x {\mathbf{x}} 
\def\y {\mathbf{y}}

In this section, we describe the proposed method in details.
% We have a scenario where there are plenty of unlabelled data, and a limited budget to annotate the examples.  
% The objective is to label within this budget the most representative examples that yield the most 
% generalisable model parameters on unseen examples. An active learning framework incrementally selects the number of training examples. From a bird's eye view, an active learning framework has three major components: a learning algorithm, a labelling oracle and a sampling/selection mechanism. 
% Given plenty of unlabelled data, at a high-level, the active learning framework has three major components: a learning algorithm, a labelling oracle and a sampling/selection mechanism. The main goal of this process is to find the most representative samples so that in a limited budget of data the highest performance is yielded.
First, we present the learners in brief for the image classification and regression tasks under the 
pool-based active learning scenario. In the second part, we discuss our two novel sampling methods: \textbf{UncertainGCN} and ~\textbf{CoreGCN}. UncertainGCN is based on the standard AL method uncertainty sampling ~\cite{Sener2017ActiveApproach} which tracks the
confidence scores of the designed graph nodes. Furthermore, CoreGCN adapts the highly successful CoreSet~\cite{Sener2017ActiveApproach} on the induced graph embeddings by the sequentially trained GCN network. 
% a novel sequential GCN selection mechanism 
% adapted with two of the best sampling techniques to date: CoreSet \cite{Sener2017ActiveApproach} and uncertainty sampling \cite{uncsamp}. We define these query methods as \textbf{CoreGCN}, the geometric approach inspired from \cite{Sener2017ActiveApproach}, and the uncertainty-based as \textbf{UncertainGCN}. A visual description of our proposed pipeline is illustrated in Figure \ref{fig:pipeline}.
%\subsection{Active Learning framework}
% \textcolor{blue}{to be continue..from here}
\subsection{Learner}

In Figure~\ref{fig:prposed_pipeline}, Phase I depicts the learner. Its goal is to 
minimise the objective of the downstream task. We have considered both the classification and regression tasks.
Thus, the objective of this learner varies with the nature of the task we are dealing with. \\
\noindent \textbf{Classification:}
For the classification tasks, the learner is a CNN image classifier.
% The learner acts as an image classifier where $\mathbf{C}$ classes are identified.
We deploy a deep model $\mathcal{M}$ that maps a set of inputs $\x \in \mathbf{X}$  to 
a discriminatory space of outputs $\y \in \mathbf{Y}$  with parameters $\textbf{$\theta$}$.
We took ResNet-18~\cite{he2016deep} as the CNN model due to its relatively higher performance 
in comparison to other networks with comparable parameter complexity.
Any other model like VGG-11\cite{Simonyan2015VeryRecognition} can also be easily deployed (refer to Supplementary Material B.4).
% We deploy a deep CNN model $\mathcal{M}$, \hspace{0.5pt} \emph{ResNet-18}~\cite{he2016deep}, 
% that maps a set of inputs $\x \in \mathbf{X}$  to 
% a discriminatory space of outputs $\y \in \mathbf{Y}$  with parameters $\textbf{$\theta$}$. Any other
% model like \emph{VGG-11}\cite{Simonyan2015VeryRecognition}
% However, as we mentioned before, any other type of discriminative model can be deployed as the objective 
% function of the learner is different from that of the sampler. 
A loss function $\mathcal{L}(\x,\y;\theta)$ is minimized during the training process. The objective 
function of our classifier is cross-entropy defined as below:
\begin{equation} \label{eq:1}
  \mathcal{L}_{\mathcal{M}}^c(\x,\y;\theta) = - \frac{1}{N_l}\sum_{i=1}^{N_l} \y_i \log(f(\x_i,\y_i;\theta)),
\end{equation}
where $N_l$ is the number of labelled training examples and $f(\x_i,\y_i;\theta)$ is the posterior probability of the model $\mathcal{M}$.

\noindent \textbf{Regression:} To tackle the 3D HPE, we deploy a well-known \textit{DeepPrior} \cite{deepprior}
architecture as model $\mathcal{M}$. Unlike the previous case, we regress the 3D hand joint coordinates from the 
hand depth images. Thus, the objective function of the model changes as in Equation~\ref{eq:reg}.
In the Equation, $J$ is the number of joints to construct the hand pose. 
% Given the main objective of regressing the 2D depth-images to 3D hand joint coordinates, the mean squared error is re-defined as the loss function accordingly:

\begin{equation} \label{eq:reg}
  \mathcal{L}_{\mathcal{M}}^r(\x,\y;\theta) =  \frac{1}{N_l}\sum_{i=1}^{N_l} \Big( \frac{1}{J}\sum_{j=1}^J\|\y_{i,j} - f(\x_{i,j},\y_{i,j};\theta)\|^2\Big),
\end{equation}
To adapt our method to any other type of task, we just need to modify the learner. 
The rest of our pipeline remains the same which we discuss in more details in the following Sections. 
% \newline

% 
% \subsection{Sequential GCN selection process}
\subsection{Sampler}
Moving to the second Phase from Figure \ref{fig:prposed_pipeline}, we adopt a pool-based scenario for active learning. 
This has become a standard in deep learning system due to its successful deployment in 
recent methods~\cite{BeluchBcai2018TheClassification,Sener2017ActiveApproach, Sinha2019VariationalLearning, Yoo2019LearningLearning}.
% pool-based scenario has become a standard in deep learning systems. 
% Suitable for deep learning models which require sets of data, we create the framework in a pool-based scenario.
In this scenario, from a pool of unlabeled dataset $\mathbf{D}_U$,
we randomly select an initial batch for labelling $\mathbf{D}_0\hspace{-1pt} \subset 
\hspace{-1pt}\mathbf{D}_U$. Without loss of generality, in active learning research, 
the major goal is to optimize a sampler's method for data acquisition, $\mathcal{A}$ 
in order to achieve minimum loss with the least number of batches $\mathbf{D}_n$. This 
scope can be simply defined for $n$ number of active learning stages as following:
\begin{equation} \label{eq:2}
  \min_n \min_{\mathcal{L}_{\mathcal{M}}} \mathcal{A}(\mathcal{L}_{\mathcal{M}}(\x,\y;\theta)|\mathbf{D}_0\hspace{-1pt}\subset\hspace{-1pt} \cdots \hspace{-1pt}\subset\hspace{-1pt} \mathbf{D}_n\hspace{-1pt}\subset \hspace{-1pt} \mathbf{D}_U).
\end{equation}
We aim to minimise the number of stages so that fewer samples $(\x,\y)$ would require annotation. For the sampling method $\mathcal{A}$, we bring the heuristic relation between the discriminative understanding of the model and the unlabelled data space. This is quantified 
by a performance evaluation metric and traced at every querying stage.

\subsubsection{Sequential GCN selection process}
\label{subsection3.2}
% In the image classification task, the input data are images $\x \in \mathbf{R}^{M\times M}$ whilst the outputs are class labels $\y \in \mathbf{C}$. Most of the CNNs architectures function in two sequential stages as a feature extractor and as a discriminator.
% The first stage is reducing the dimensions of the data space into shared spatial representations. The second part is linearly regressing to the fixed $\mathbf{C}$ classes. Given the initial labelled set $\mathbf{D}_0$, we train our learner to evaluate the predicted classes. 
During sampling as shown in Figure~\ref{fig:prposed_pipeline} from Phase II to IV,  our contribution relies on sequentially training a GCN initialised with the features extracted from the learner for both labelled and unlabelled images at every active learning stage. As stated before, similar to VAAL \cite{Sinha2019VariationalLearning}, we consider this methodology as model-based  where a separate architecture is required for sampling. Our motivation in introducing the graph is primarily in propagating the inherited uncertainty on the learner feature space between the samples (nodes). Thus, message-passing between the nodes induces higher-order representations of the nodes after applying convolution on the graph. Finally, our GCN will act as a binary classifier deciding which images are annotated.
\def\G {\mathcal{G}}
\def\V {\mathcal{V}}
\def\E {\mathcal{E}}
\def\v {\mathbf{v}}

\noindent \textbf{Graph Convolutional Network.} The key components of a graph, $\G$ are the nodes, also called vertices $\V$ and the edges $\E$. The edges capture the
relationship between the nodes and encoded in an adjacency matrix $A$. The nodes $\v \in \mathbb{R}^{(m \times N)}$ of the graph encode image-specific information and are initialised with the features extracted from the learner. Here, ~$N$ represents the total number of both labelled and unlabelled examples while 
$m$ represents the dimension of features for each node.
After we apply $l_2$ normalisation to the features, the initial elements of $A$ result as vector product between every sample of $\v$ i.e. ($S_{ij}=\v_i^{\top} \v_j, \{i, j\} \in N$). This propagates the similarities between nodes while falling under the same metric space as the learner's objective. Furthermore, we subtract from $S$ the identity matrix $I$ and then we normalise by multiplying with its degree $D$. Finally, we add the self-connections back so that the closest correlation is with the node itself. This can simply be summarised under:
\begin{equation} \label{eq:4}
    A = D^{-1}(S-I)+I.
\end{equation}

To avoid over-smoothing of the features in GCN \cite{kipf2016semi}, we adopt a two-layer architecture. The first GCN layer can be described as a function $f_\G^1(A, \V; \Theta_1):\mathbb{R}^{N\times N} \times \mathbb{R}^{m\times N} \rightarrow \mathbb{R}^{h\times N}$ where $h$ is number of hidden units and $\Theta_1$ are its parameters. A rectified linear unit activation \cite{relu} is applied after the first layer to maximise feature contribution. However, to map the nodes as labelled or unlabelled, the final layer is activated through a sigmoid function. Thus, the output of $f_{\G}$ is a vector length of $N$ with values between 0 and 1 (where 0 is considered unlabelled and 1 is for labelled). We can further define the entire network function as:
\begin{equation}
    f_{\G}= \sigma(\Theta_2(ReLU(\Theta_1A)A).
\end{equation}
In order to satisfy this objective, our loss function will be defined as:

\begin{equation}
    \begin{split}
\mathcal{L}_{\mathcal{G}}(\V,A;\Theta_1,\Theta_2) &= - \frac{1}{N_l}\sum_{i=1}^{N_l} \log(f_{\G}(\V,A;\Theta_1,\Theta_2)_i) -\\ 
 & \hspace{-28pt}- \frac{\lambda}{N-N_l}\sum_{i=N_l+1}^N\log(1-f_{\G}(\V,A;\Theta_1,\Theta_2)_i),
      \end{split}
\end{equation}
where $\lambda$ acts as a weighting factor between the labelled and unlabelled cross-entropy.
\\
% \subsubsection{UncertainGCN: Uncertainty sampling on GCN}
\noindent \textbf{UncertainGCN: Uncertainty sampling on GCN.} Once the training of the GCN is complete, we move forward to selection. From the remaining unlabelled samples $\mathbf{D}_U$, we can draw their confidence scores $f_{\G}(\v_i ;  \mathbf{D}_U)$ as outputs of the GCN. Similarly to uncertainty sampling, we propose to select with our method, \textbf{UncertainGCN}, the unlabelled images with the confidence depending on a variable $s_{margin}$. While querying a fixed number of $b$ points for a new subset $\mathbf{D}_L$, we apply the following equation:
\begin{equation} \label{eq:3}
    \mathbf{D}_L = \mathbf{D}_L  \cup \argmax_{i=1\cdots b}\lvert s_{margin}- f_{\G}(\v_i ;  \mathbf{D}_U\hspace{-1pt})\rvert.
\end{equation}

For selecting the most uncertain unlabelled samples, $s_{margin}$ should be closer to 0. In this manner, the selected images are challenging to discriminate, similarly to the adversarial training scenario \cite{goodfellow2014generative}. 
This stage is repeated as long as equation \ref{eq:2} is satisfied. Algorithm \ref{alg:1} summarises the GCN sequential training with the UncertainGCN sampling method.
 
 \begin{algorithm} 
   \caption{UncertainGCN active learning algorithm}
   \begin{algorithmic}[1]
      \State \textbf{Given}: Initial labelled set $\mathbf{D}_0$, unlabelled set $\mathbf{D}_U$ and query budget $b$
      \State \textbf{Initialise} $(\x_L, \y_L) ,(\x_U)$ - labelled and unlabelled images
      \Repeat
      
          \State $\theta \leftarrow f(\x_L, \y_L)$ \Comment{Train learner with labelled}
          \State $\V = [\v_L, \v_U] \leftarrow f{(\x_L \cup \x_U ; \theta)}$ \Comment{Extract features for labelled and unlabelled}
          \State \textit{Compute adjacency matrix $A$ according to Equation \ref{eq:4}}
          \State $\Theta \leftarrow f_\G(\V, A)$ \Comment{Train the GCN}
   
          \For{$i = 1 \to b$}
              \State $\mathbf{D}_L$ = $\mathbf{D}_L \cup \argmax_i \lvert s_{margin}- f_{\G}(\v ;  \mathbf{D}_U)\rvert$ 
              \Comment{Add nodes depending on the label confidence}
          \EndFor
          \State \textit{Label $\y_U$ given new $\mathbf{D}_L$}
          
      \Until {\textit{Equation \ref{eq:2}} is satisfied}
\end{algorithmic}
\label{alg:1}
\end{algorithm}
% \\
% \subsubsection{CoreGCN: CoreSet sampling on GCN.}
\noindent \textbf{CoreGCN: CoreSet sampling on GCN.}
\label{subsection3.3}
To integrate geometric information between the labelled and unlabelled graph representation, we approach a CoreSet technique \cite{Sener2017ActiveApproach} in our sampling stage. This has shown better performance in comparison to uncertainty-based methods \cite{Wu2019ActivePropagation}.  \cite{Sener2017ActiveApproach} shows how bounding the difference between the loss of the unlabelled samples and the one of the labelled is similar to the \emph{k-Centre} minimisation problem stated in \cite{wolf}. 

In this approach, the sampling is based on the $l2$ distances between the features extracted from the trained classifier.
Instead of that, we will make use of our GCN architecture by applying CoreSet method on the features represented after the first layer of the graph. To this, the CoreSet method benefits from the cyclical dependencies. The sampling method is adapted to our mechanism for each $b$ data point under the equation:
\begin{equation}
    \mathbf{D}_L = \mathbf{D}_L  \cup \argmax_{i \in \mathbf{D}_U} \min_{j \in \mathbf{D}_L} \delta( f_\G^1(A, \v_i; \Theta_1), f_\G^1(A, \v_j; \Theta_1)),
\end{equation}
where $\delta$ is the Euclidean distance between the graph features of the labelled node $\v_i$ and the ones from the unlabelled node $ \v_j$.  We define this method as \textbf{CoreGCN}.

Finally, given the model-based mechanism, we claim that our sampler is task-agnostic as long as the learner is producing a form of feature representations. In the following section, we will experimentally demonstrate the performance of our methods quantitatively and qualitatively.
% \label{headings}

% \section{Experiments}
\section{Experiments}
\label{experiments}
We performed experiments on sub-sampling RGB and grayscale real images for classification,
depth real images for regression and RGB synthetic-generated for classification tasks. We describe them in details below.
\subsection{Classification}
\noindent \textbf{Datasets and Experimental Settings.} 
We evaluated the proposed AL methods on four challenging image classification 
benchmarks. These include three RGB image datasets, CIFAR-10\cite{cifar}, CIFAR-100\cite{cifar} and SVHN\cite{Goodfellow2014Multi-digitNetworks},
and a grayscale dataset, FashionMNIST\cite{Xiao2017Fashion-MNIST:Algorithms}.
Initially, for every benchmark, we consider the entire training set as an unlabelled pool ($\mathbf{D}_U$). As a cold-start, we randomly sample a small subset and query their labels, $\mathbf{D}_L$. 
For CIFAR-10, SVHN and FashionMNIST, the size of the seed labelled examples is 1,000. Whereas, for CIFAR-100 we select 2,000
due to their comparatively more number of classes (100 vs 10). 
We conduct our experiments for 10 cycles. At every stage, the budget is fixed at
1,000 images for the 10-class benchmarks and at 2,000
for CIFAR-100 which is a 100-class benchmark. 
Similar to the existing works of \cite{BeluchBcai2018TheClassification,Yoo2019LearningLearning},
we apply our selection on randomly selected subsets $\mathbf{D}_S \hspace{-1pt}\subset\hspace{-1pt} \mathbf{D}_U$ 
of unlabelled images. This avoids the redundant occurrences which are common in all datasets~\cite{data_redundant}. The size of $\mathbf{D}_S$ is set to 10,000 for all the experiments.
% The only difference for the synthetic experiment is that the model starts pre-trained to select  example is we don't select seed labelled examples since we are interested in selecting quality examples.
\\
% \noindent \textbf{Implementation details.} 
\noindent \textbf{Implementation details.} 
ResNet-18 \cite{he2016deep}  is the favourite choice as learner due to its relatively  
higher accuracy and better training stability.
% We also compared performance on VGG-11~\cite{Simonyan2015VeryRecognition}.
During training the learner, we set a batch size of 128. We use Stochastic Gradient Descent (SGD) 
with a weight decay $5 \times 10^{-4}$ and a momentum of $0.9$. At every selection stage, we train the model for
200 epochs. We set the initial learning rate of 0.1 and decrease it by the factor of 10 after 160 epochs. 
We use the same set of hyper-parameters in all the experiments. 
% The first part of our selection mechanism is approximated by the GCN. 
For the sampler, GCN has 2 layers and we set the dropout rate to 0.3 to avoid over-smoothing~\cite{zhao2019pairnorm}.
The dimension of initial representations of a node is 1024 and it is projected to 512.
The objective function is binary cross-entropy per node. 
We set the value of $\lambda = 1.2$ to give more importance to the larger unlabelled subset. 
We choose Adam \cite{Kingma2015ADAM:Optimization} optimizer with a weight decay of $5 \times 10^{-4}$ and 
a learning rate of $10^{-3}$. We initialise the nodes of the graph with the features of the images extracted from the learner. 
% This initialisation enables the sampler to understand the need for learner's complimentary examples.
% \textcolor{green}{The dimension of the features projected to the GCN is 512}
 We set the value of $s_{margin}$ to 0.1. For the empirical comparisons, we suggest readers to refer Supplementary Material.
%  This value matches with our expectation as we are selecting unlabelled examples sufficiently different/further than already labelled ones {This should go to cross-validation section}.
% $s_{margin}$ variable to set.  From our ablation studies, we observed that our method selects the most informative samples when $s_{margin}$ equals 0.1. This is intuitively correct as unlabelled samples with scores further from 0.1 will be selected. } \\
% \noindent \textbf{Compared methods.} We compare the performance of our i
\\
\noindent \textbf{Compared Methods and Evaluation Metric:} We compare our method with
a wide range of baselines which we describe here. 
Random sampling is by default the most common sampling technique. 
CoreSet\cite{Sener2017ActiveApproach} on learner feature space is one of the
best performing geometric techniques to date and it is another competitive baseline for us. 
VAAL~\cite{Sinha2019VariationalLearning} and Learning Loss~\cite{Yoo2019LearningLearning} 
are two state-of-the-art baselines from task-agnostic frameworks.
Finally, we also compare with FeatProp~\cite{Wu2019ActivePropagation} which 
is a representative baseline for the GCN-based frameworks. This method is designed for cases 
where a static fixed graph is available. To approximate their performance, we construct 
a graph from the features extracted from learner and similarities between the features
as edges. We then compute the k-Medoids distance on this graph.
For quantitative evaluation, we report the mean average accuracy of 
5 trials on the test sets.

% \textcolor{green}{ The second baseline is
% CoreSet\cite{Sener2017ActiveApproach} which
%     % The purpose of this method is to 
%     finds $b$ unlabelled samples that will act as new
%     centres while the now reduced radius will cover the entire feature space from the
%     learner. VAAL~\cite{Sinha2019VariationalLearning} is another method to compare with. We consider this work as one of the closest to ours since this trains VAE with both labelled and unlabelled samples to learn a discriminative subspace in 
%     an adversarial manner.} Learning Loss~\cite{Yoo2019LearningLearning} is another state-of-the-art method which tries to minimise an additional loss to predict the loss
%     in a self-supervised fashion. Finally, we compared the performance of FeatProp~\cite{Wu2019ActivePropagation} with our method. This method is state-of-the-art for 
%     GCN-based samplers. \textcolor{green}{As briefly mentioned in Section~\ref{sec:intro},
%     FeatProp, designed for a GCN learner, relies on predefined features of both labelled and unlabelled. Furthermore, similar to CoreSet, it approaches for selection a geometric computation of k-Medoids based on edge distances. To adapt it for a different type of learner, we pass the features as nodes, assign the graph and evaluate the k-Medoids distances.} For quantitative evaluation, we report the mean average accuracy of the 5 trials on test sets. 
\begin{figure}
    \centering
    \includegraphics[trim=2.5cm 0cm 3.5cm 1.0cm, clip, width=0.80\linewidth]{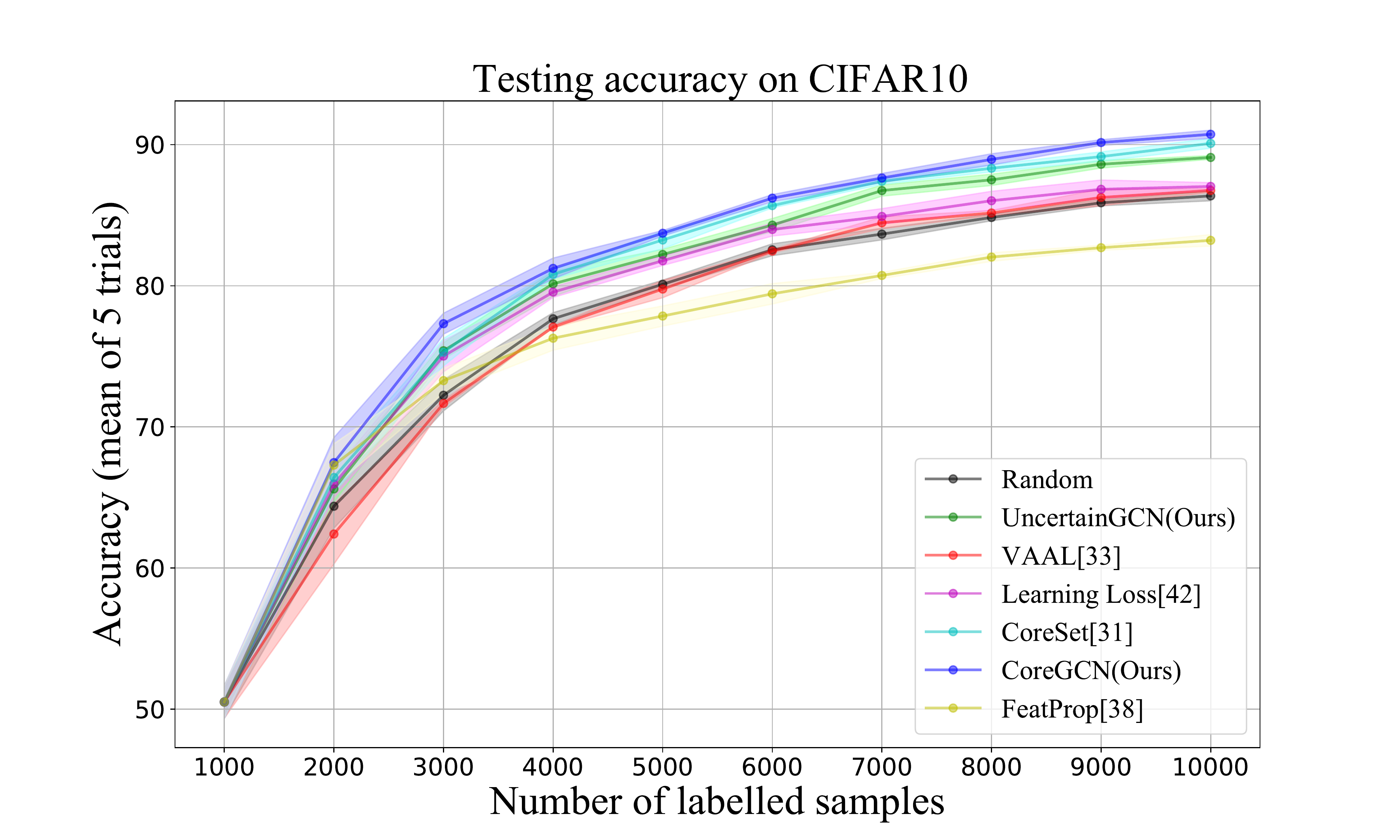}
    \includegraphics[trim=2.5cm 0cm 3.5cm 1.0cm, clip, width=0.80\linewidth]{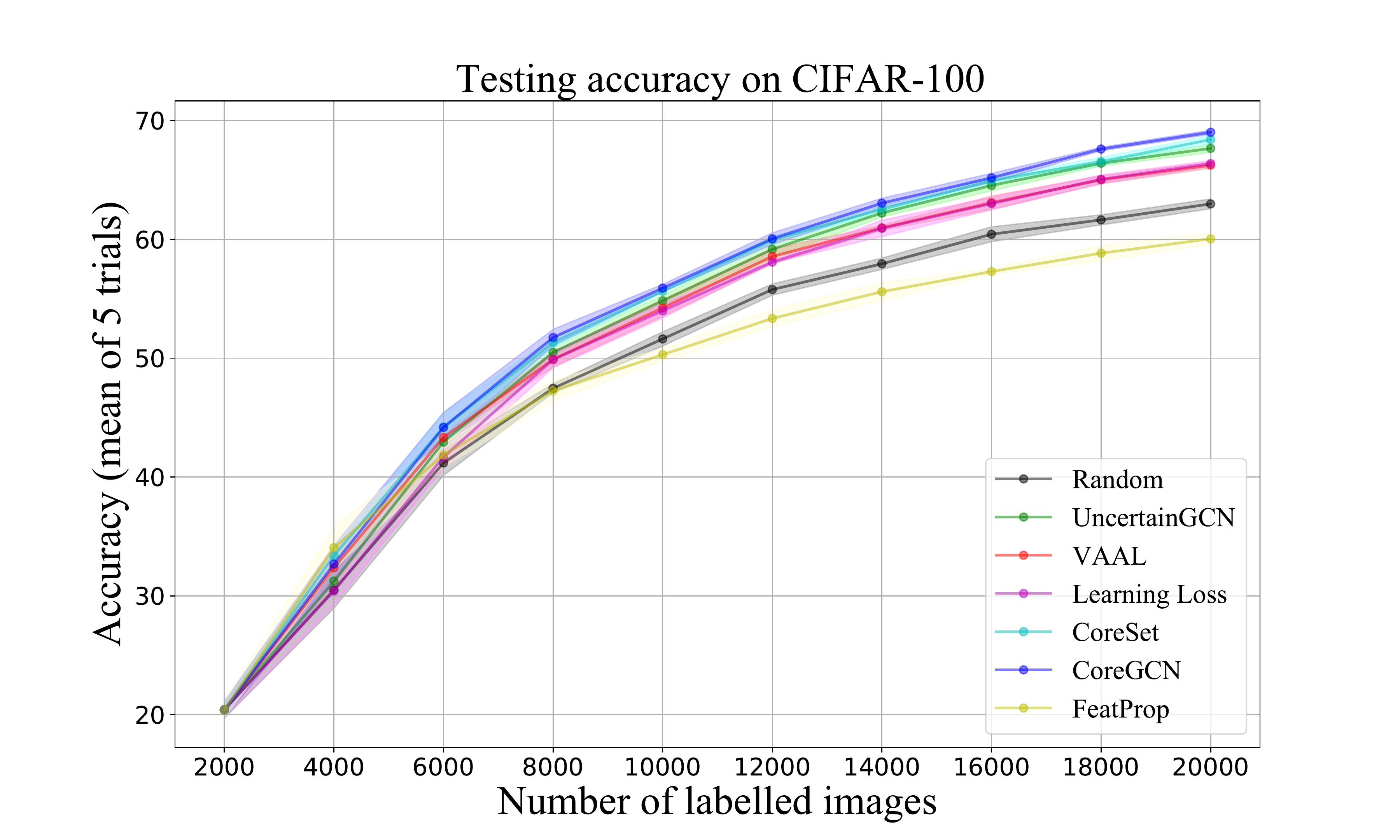}
    \caption{Quantitative comparison on CIFAR-10(top) and CIFAR-100(bottom) (Zoom in the view)}
    \label{fig:cf10}
\end{figure}
\begin{figure}
    \centering
    \includegraphics[trim=2.5cm 0cm 3.5cm 1cm, clip, width=0.80\linewidth]{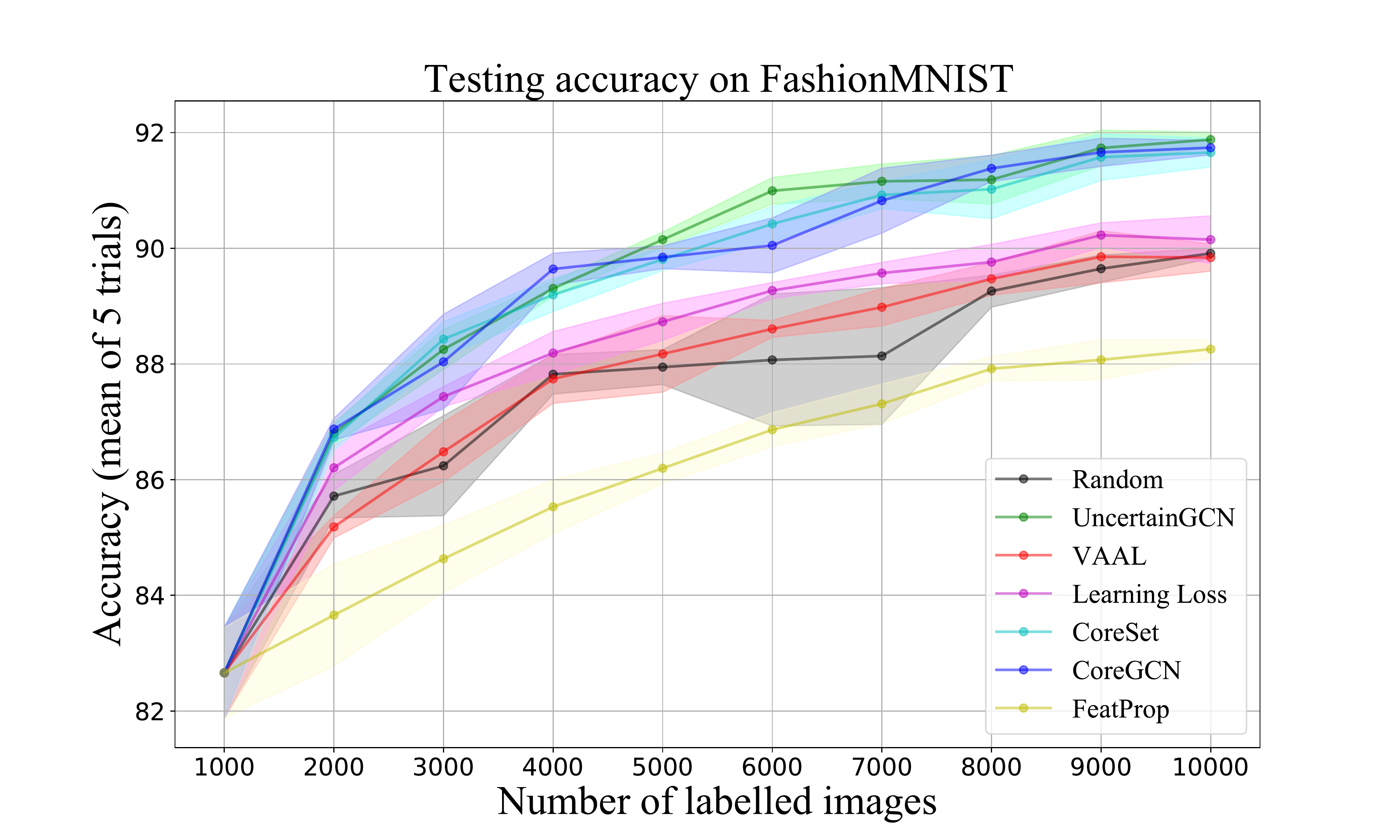}
    \includegraphics[trim=2.5cm 0cm 3.5cm 1cm, clip,  width=0.80\linewidth]{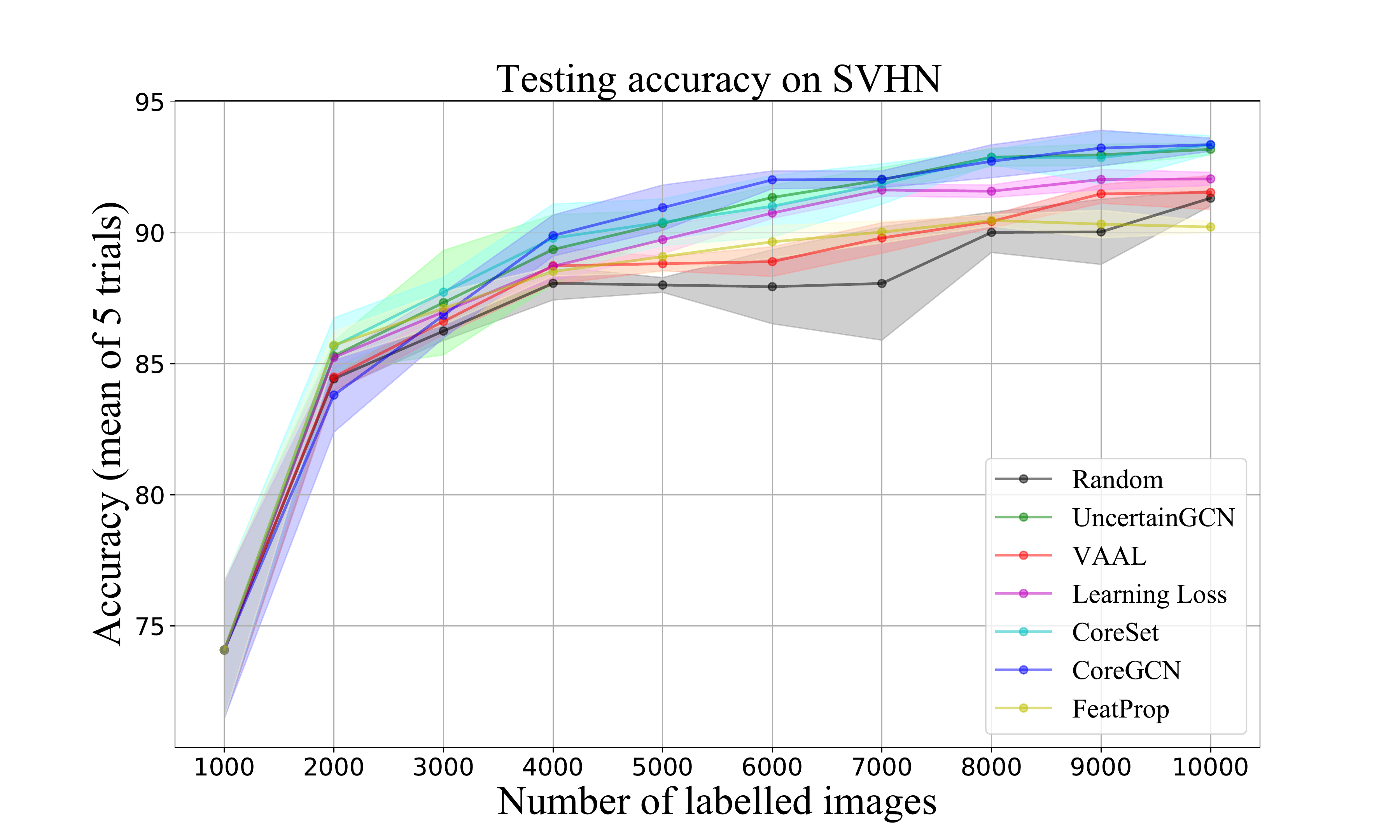}
    \caption{Quantitative comparison on FashionMNIST(top) and SVHN(bottom) (Zoom in the view)}
    \label{fig:fm}
\end{figure}
\noindent \textbf{Quantitative Comparisons.} We train the ResNet-18 learner with all the available training examples on every dataset separately and report the performance on the test set. Our 
implementation obtains 93.09\% on CIFAR-10, 73.02\% on CIFAR-100, 93.74\% on FashionMNIST, 
and  95.35\% on SVHN. This is comparable as reported on the official implementation~\cite{he2016deep}.
% \textcolor{red}{cite if there are any} and this checks the sanity of our implementation. 
These results are also set as the upper-bound performance of the active learning frameworks.
% Discussion on numbers for each dataset. What is the performance of the model with the entire dataset? what is the balance between the amount of data to annotate and accuracy?

Figure \ref{fig:cf10} (left) shows the performance comparison of UncertainGCN and CoreGCN 
with the other five existing methods on \textbf{CIFAR-10}. The solid line of the representation 
is the mean accuracy and the faded colour shows the
standard deviation. Both our sampling techniques surpass almost every other compared methods in every selection stage.
CoreSet is the closest competitor for our methods. After selecting 10,000 labelled examples, 
the CoreGCN achieves 90.7\% which is the highest performance amongst reported in the literature~\cite{Yoo2019LearningLearning,Sinha2019VariationalLearning}. 
% The performance obtained with a fifth of $\mathbf{D}_U$ brings an important factor to the research community in whether labelling 40,000 more images is needed for a 2.4\% gain.
% \textcolor{green}{ UncertainGCN comes close to CoreSet achievements over the testing set. However, in latter selection stages its performance is restricted due to the over-selection in uncertain feature-space areas. This aspect is demonstrated in the qualitative analysis.}
%Cifar100 
Likewise, Figure \ref{fig:cf10}~(right) shows the accuracy comparison on \textbf{CIFAR-100}. 
We observe almost similar trends as on CIFAR-10.  
% we indicate that our proposed methods can scale to a more diverse dataset maintaining their state-of-the-art results. 
With only 40\% of the training data, we achieve 69\% accuracy by applying CoreGCN.
This performance is just 4\% lesser than when training with the entire dataset. 
Compared to CIFAR-10, we observe the better performance on VAAL 
in this benchmark.
% shows a better trend against random sampling, but it still falls under our performances.
The reason is that VAE might favour a larger query batch size ($>$1,000).
This exhaustively annotates large batches of data when the purpose of active learning 
is to find a good balance between exploration and exploitation as we constrain
the budget and batches sizes.

%FashionMNST and SVHN
We further continue our evaluation on the image classification by applying our methods on 
FashionMNIST and SVHN. In Figure \ref{fig:fm}, the left and the right graphs show the comparisons on
\textbf{FashionMNIST} and \textbf{SVHN} respectively. As in the previous cases, our methods achieve at minimum 
similar performance to that of existing methods or outperforming them.
From the studies on these datasets, we observed consistent modest performance of FeatProp \cite{Wu2019ActivePropagation}.
This may be because it could not generalise on the unstructured 
data like ours.
% The reason stands on the learner's design constraint to a Graph Convolutional Network. Hence, this is not suitable for the cases where the image classifier is different from the sampler. \textcolor{green}{The accuracy shows promising results for the first selection stages, but then it gains minimal improvement on CIFAR-10, CIFAR-100 and FashionMNIST, falling behind random sampling. A similar plateauing effect can be noticed in \cite{Wu2019ActivePropagation} on the PubMed dataset... What is not clear? }
% Overall comparisons demonstrate the clear superior performance of the proposed methods compared to the recent state-of-the-art in active learning.
\begin{figure}
    \centering
    \includegraphics[trim=0cm 0cm 0cm 0cm, clip, width=0.49\linewidth]{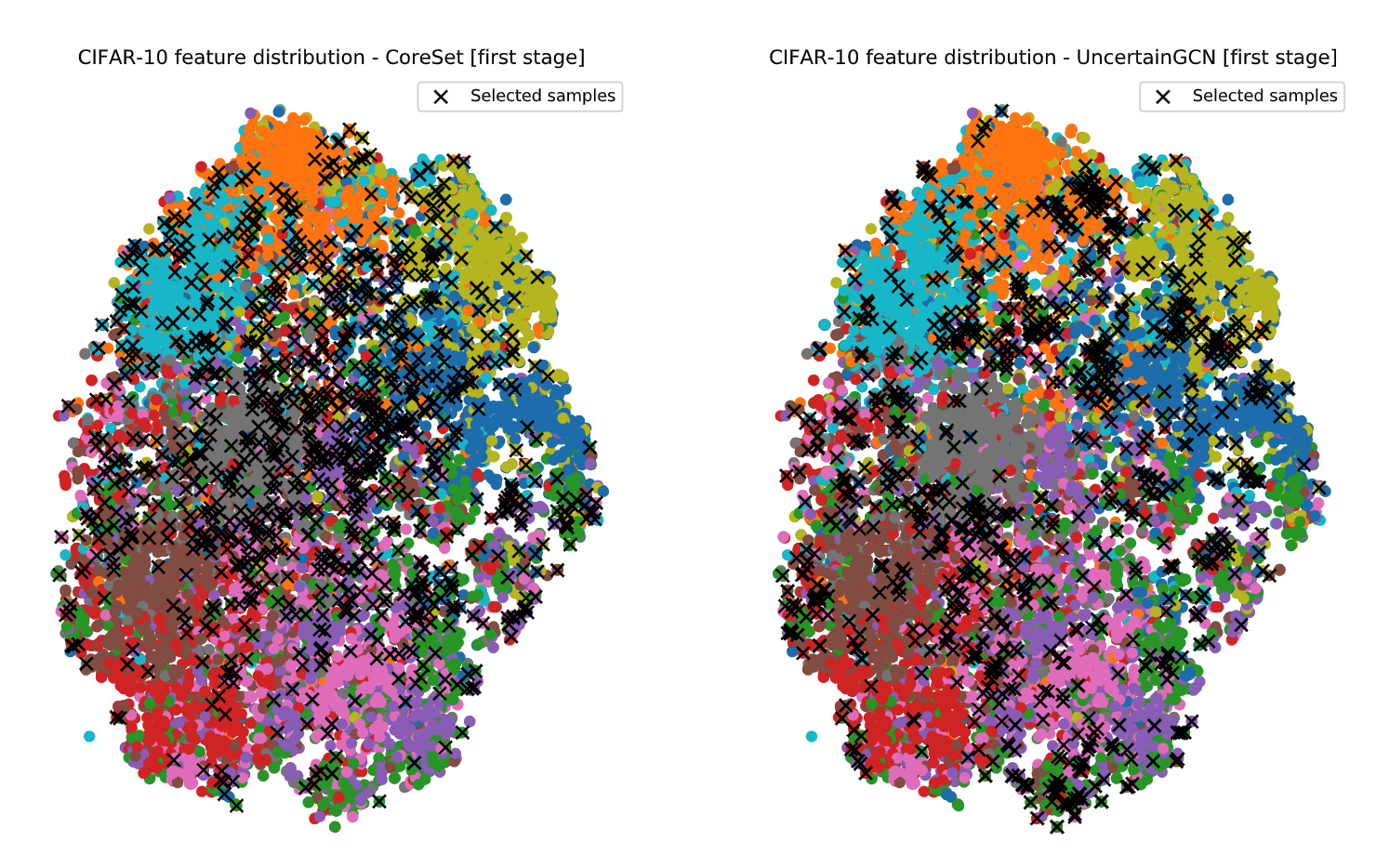}
    \includegraphics[trim=0cm 0cm 0cm 0cm, clip,  width=0.49\linewidth]{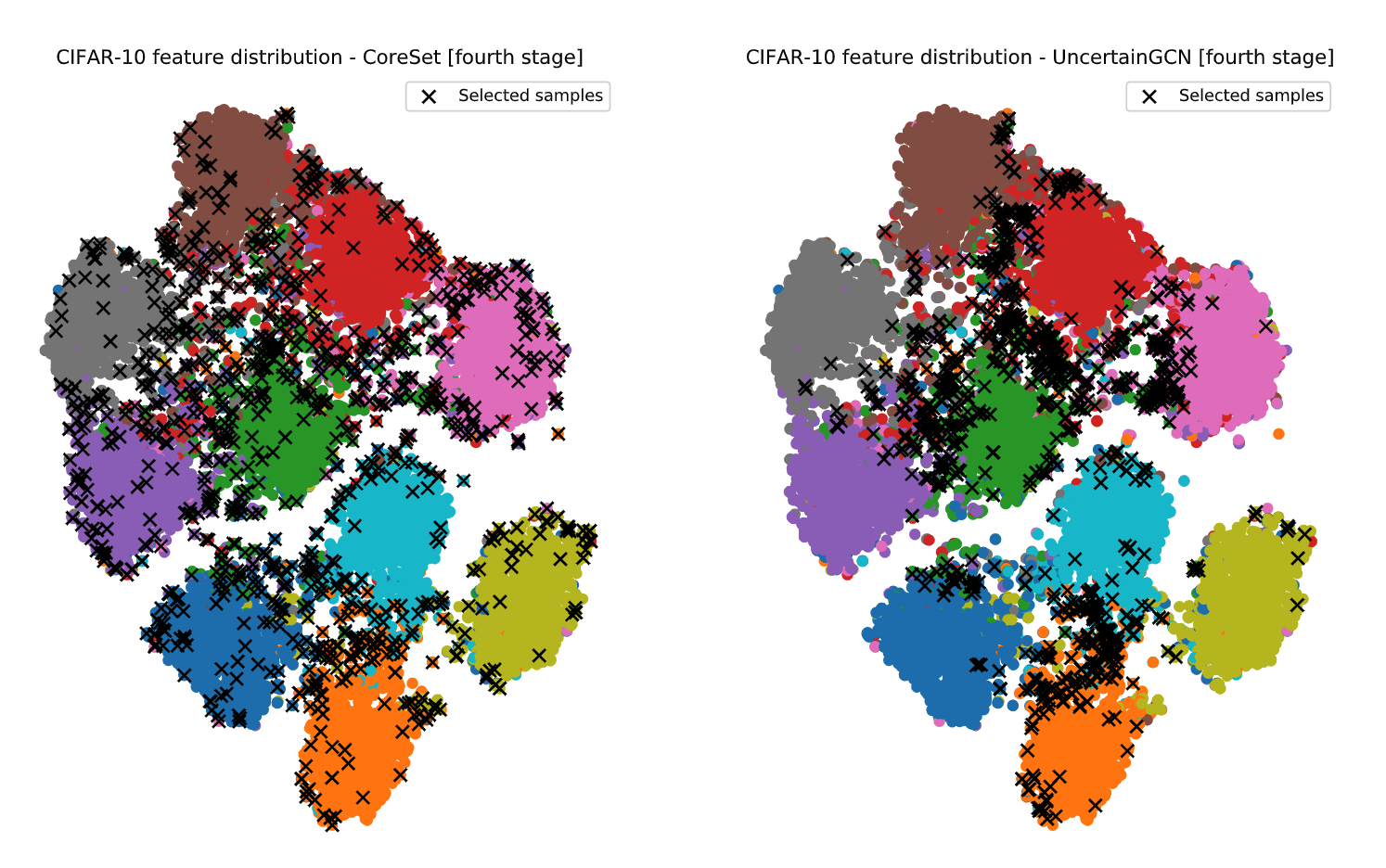}
    \caption{Exploration comparison on CIFAR-10 between CoreSet and UncertainGCN (Zoom in the view)}
    \label{fig:tsne}
\end{figure}

\begin{figure}
    \centering
    \includegraphics[trim=0cm 0cm 0cm 0cm, clip, width=0.49\linewidth]{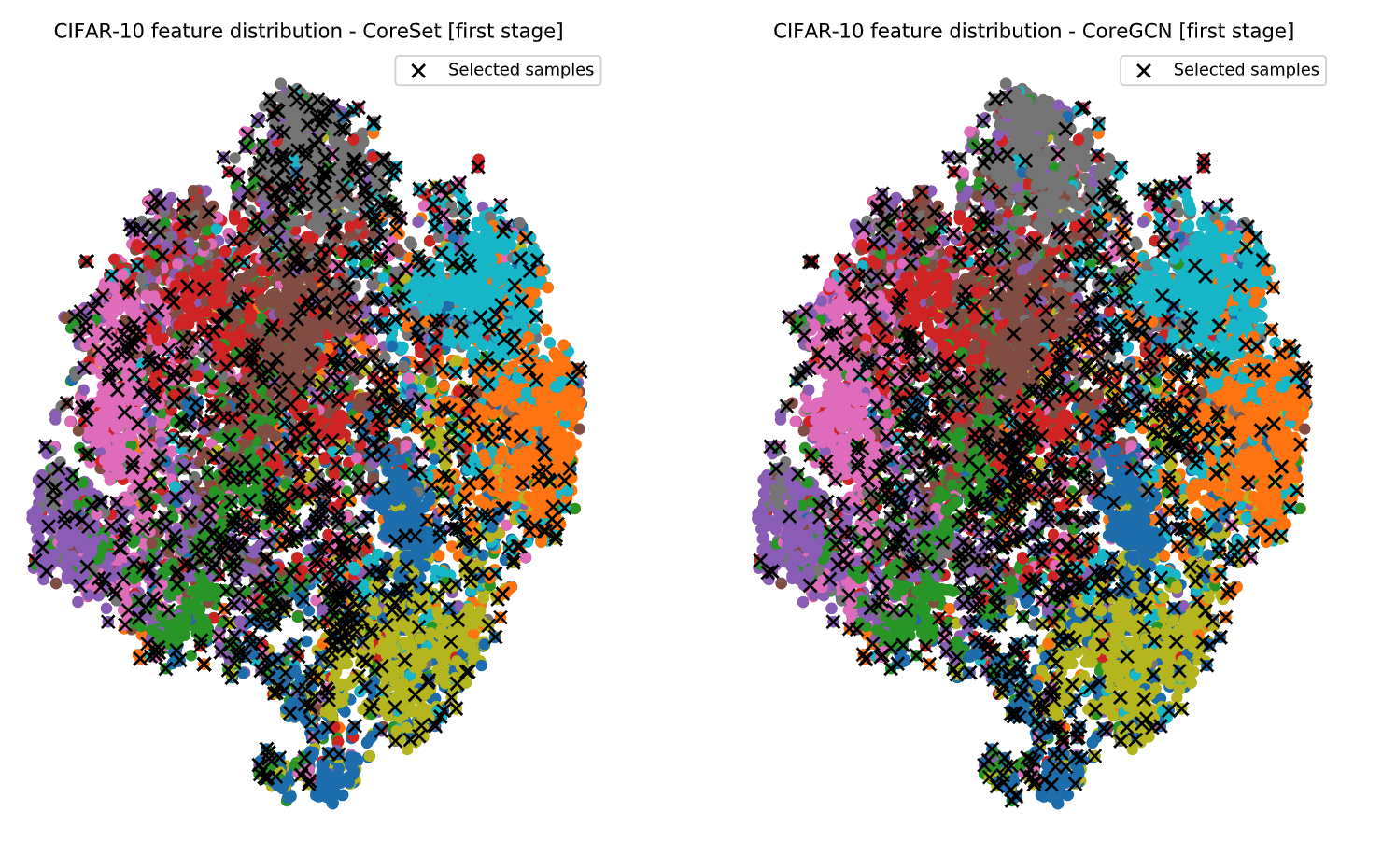}
    \includegraphics[trim=0cm 0cm 0cm 0cm, clip,  width=0.49\linewidth]{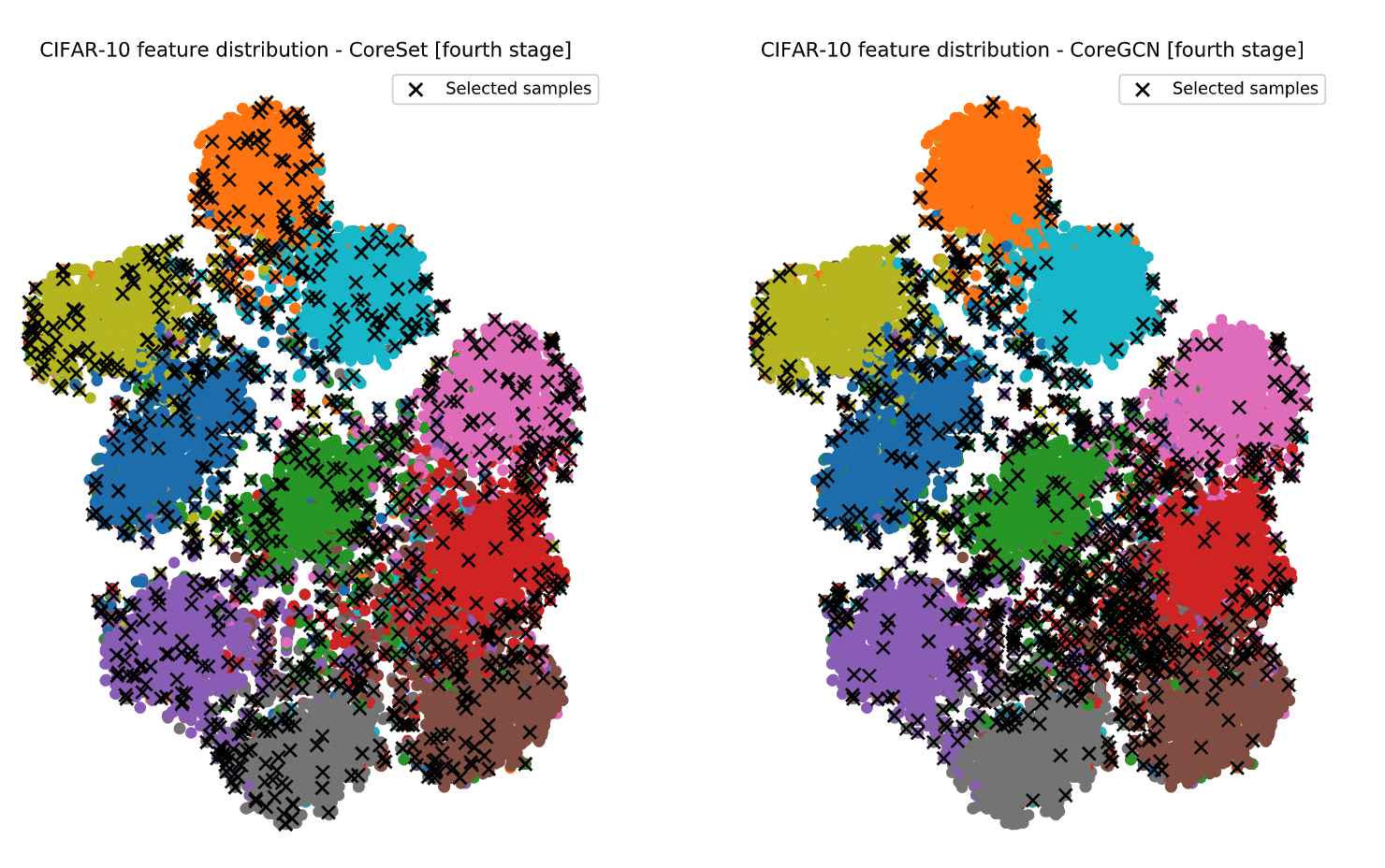}
    \caption{Exploration comparison on CIFAR-10 between CoreSet and CoreGCN (Zoom in the view)}
    \label{fig:tsne_compare_coreset}
\end{figure}

\noindent \textbf{Qualitative Comparisons.}
%  In active learning, for image classification, it is convenient to elaborate a qualitative comparison through distributions in the data space.
To further analyse the sampling behaviour of our method we perform qualitative comparison with 
existing method. We choose CoreSet for its consistently better performance in 
empirical evaluations when compared to the other baselines.
% we deploy a qualitative analysis too. Therefore, we choose CoreSet to represent the baselines 
% and UncertainGCN to represent our contribution. 
We made this comparison on CIFAR-10.
For the two algorithms, we generate the t-SNE \cite{Maaten08visualizingdata} plots of 
both labelled and unlabelled extracted features from the learner at 
the first and fourth selection stage. To make a distinctive comparison of sampling behaviour 
from early stage, we choose to keep a difference 
of 3 stages. 
Figure~\ref{fig:tsne}, t-SNE plots, compares the sampling behaviour of CoreSet and UncertainGCN.
% We can notice that each cluster represents a category of the images.
In the first sampling stage, the selected samples distribution
is uniform which is similar for both techniques. 
Without loss of generality, the learner trained with a small 
number of seed annotated examples is sub-optimal, and, hence 
the features of both labelled and unlabelled are not discriminative enough.
This makes the sampling behaviour for both methods near random. 
% This is depicted by the uniform distribution of the selected  example in the first t-SNE. 
At the fourth selection stage, the learner becomes relatively more discriminative. 
This we can notice from the clusters representing each class of CIFAR-10.
Now, these features are robust to capture the relationship between the 
labelled and unlabelled examples which we encode in the adjacency matrix. 
Message-passing operations on GCN exploits the correlation between 
the labelled and unlabelled examples by inducing similar representations.
This enables our method to target on the out-of-distribution
unlabelled samples and areas where features are hardly distinguished. This characteristics we can observe 
on the plot of the fourth selection stage of UncertainGCN. 
% This happensdue to querying the most uncertain images where the unlabelled are 
% confused with labelled. 
Similarly, in Figure \ref{fig:tsne_compare_coreset}, 
we continue the qualitative investigation for the CoreGCN acquisition method.
CoreGCN avoids over-populated areas while tracking out-of-distribution unlabelled data.
Compared to UncertainGCN, the geometric information from CoreGCN maintains a 
sparsity throughout all the selection stages. Consequently, it preserves
the message passing through the uncertain areas while CoreSet keeps sampling 
closer to cluster centres. This brings a stronger balance in comparison to 
CoreSet between in and out-of-distribution selection with the availability of
more samples.

\begin{figure}
    \centering
    \includegraphics[trim=2cm 0cm 1cm 1cm, clip, width=0.80\linewidth]{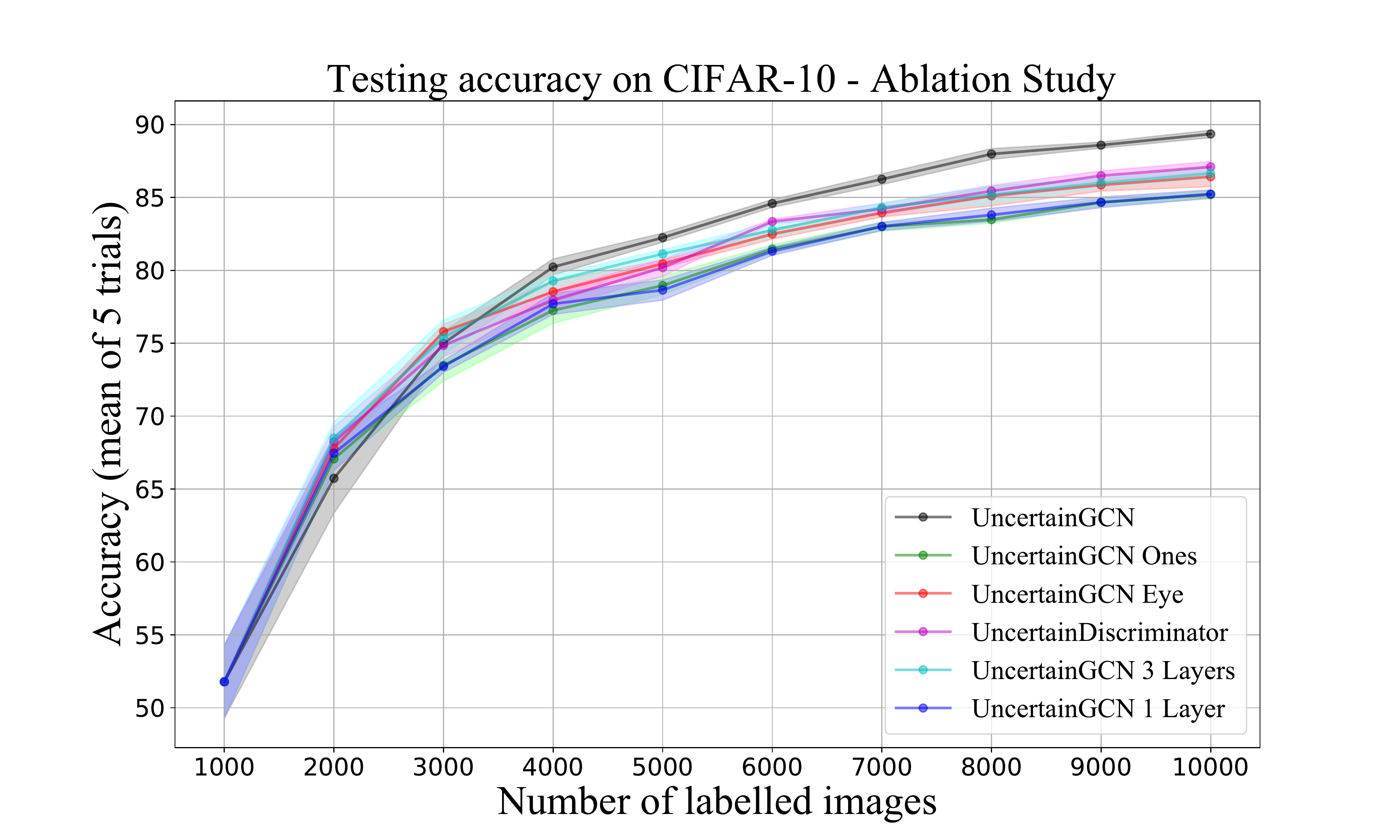}
    % \caption{Quantitative evaluations: Our method (UncertainGCN), adjacency matrix fixed filled with 1s (UncertainGCN Ones), Identity matrix as adjacency matrix (UncertainGCN Eye), Labelled/Unlabelled Discriminator on learner feature space (UncertainDriscrimator), UncertainGCN with a single and 3-layer GCN (Please zoom in)}
    \caption{Ablation studies (Please zoom in)}
    \label{fig:abl_st}
\end{figure}

\noindent \textbf{Ablation Studies} To further motivate the GCN proposal, we conduct ablation studies on the sampler architecture. In Figure ~\ref{fig:abl_st}, still on CIFAR-10, we replace the GCN with a 2 Dense layer discriminator, \emph{UncertainDiscriminator}. This approach over-fits at early selection stages.
Although, GCN with 2 layers \cite{kipf2016semi} has been a de-facto optimal design choice,
we also report the performance with 1 layer (hinders long-range propagation) and 3 (over-smooths). However, to further quantify the significance of our adjacency matrix with feature correlations, we evaluate GCN samplers with identity (UncertainGCN Eye) and filled with 1s matrices (UncertainGCN Ones).
Finally, a study on 
% we report the performance on different combinations of 
two important hyper-parameters: drop-out (0.3, 0.5, 0.8) and the number of hidden units (128, 256, 512) is in the Supplementary B.2. 
We also fine-tune these parameters to obtain the optimal solution.

\subsection{Regression} 
\noindent \textbf{Dataset and Experimental Settings:} We further applied our method on 
a challenging dataset for 3D Hand Pose Estimation benchmarks from depth images. 
ICVL~\cite{icvl} contains 16,004 hand depth-images in the training set and the test set has 1,600. At every selection stage, similar to the experimental setup of image classification, we 
randomly pre-sample 10\% of entire training examples $\mathbf{D}_S$ and apply the AL methods on 
this subset of the data. Out of this pre-sampled images subset, we apply our sampler to select the 
most influencing 100 examples.
% Furthermore, we choose a budget of 1000 examples for NYU and BigHand2.2M, whereas, we keep only 100
% for ICVL. This is because the ICVL benchmark 
% Similarly to classification experiments, 
% we run over 5 times, while we randomly sub-sample to $\mathbf{D}_S$ before 
% applying the AL method. Specifically, we consider subsets of 10\% from 
% the entire training set. The budgets for NYU and BigHand2.2M  
% are limited to 1000 samples while for ICVL is 100. 
% This has been adapted due to each dataset size
\\
\noindent \textbf{Implementation Details:}
3D HPE is a regression problem  which involves estimating the 3D coordinates of the hand joints from depth images.
Thus, we replace ResNet-18 by commonly used \emph{DeepPrior} \cite{deepprior} as learner.
The sampler and the other components in our pipeline remain the same as in 
the image classification task. This is yet another evidence for our
sampling method being task-agnostic. For all the experiments, we train the 3D HPE with
Adam\cite{Kingma2015ADAM:Optimization} optimizer and with a learning rate of $10^{-3}$.
The batch size is 128. As pre-processing, we apply a pre-trained U-Net ~\cite{unet} model to detect hands, 
centre, crop and resize images to the dimension of 128x128.
% As in prior works,
% we pre-process the images by centring and cropping with a hand detection algorithm based on
% U-Net\cite{unet}. 
\\
\noindent \textbf{Compared Methods and Evaluation Metric:}
We compare our methods from the two ends of the spectrum of baselines. One is random sampling 
which is the default mechanism. The other is CoreSet\cite{Sener2017ActiveApproach}, one of the best 
performing baselines from the previous experiments.
% In our image classification experiments, we also observed 
% a trend of CoreSet outperforming the rest of the compared baselines.
We report the performance in terms of mean squared error averaged from 5 different trials 
and its standard deviation. 
% as baselines because the latter has shown greater performance in image classification than 
% the other methods
% VAAL\cite{Sinha2019VariationalLearning}, Learning Loss\cite{Yoo2019LearningLearning}, and FeatProp\cite{Wu2019ActivePropagation}. 
\begin{figure}%
    \centering
    \includegraphics[trim=2.5cm 0.2cm 3.5cm 1cm, clip,  width=.75\linewidth]{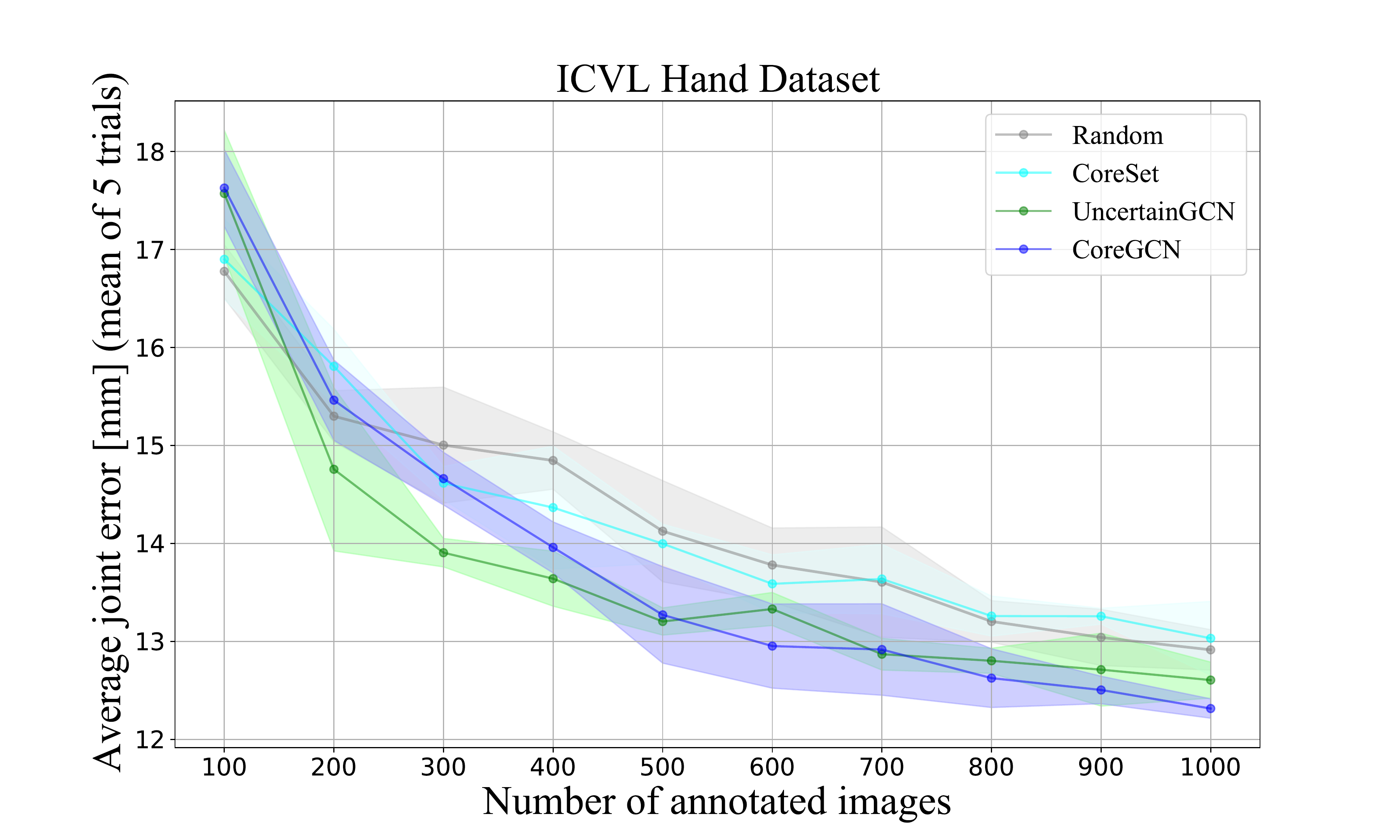}
    \caption{Quantitative comparison on 3D Hand Pose Estimation (lower is better)}%
    \label{fig:hands}%
\end{figure}
\\
\noindent \textbf{Quantitative Evaluations:} Figure~\ref{fig:hands} shows the performance comparison 
on ICVL dataset. In the Figure, we can observe that both our sampling methods, CoreGCN and UncertainGCN,
outperform the CoreSet and Random sampling consistently from the second selection stage.
The slope of decrease in error for our methods sharply falls down from the second 
till the fifth selection stage for UncertainGCN and till the sixth for CoreGCN. 
This gives us an advantage over the other methods when we have a very limited budget. 
At the end of the selection stage, CoreGCN gives the least error of 12.3 mm. 
In terms of performance, next to it is UncertainGCN. 
% another our method UncertainGCN. 
% \textcolor{blue}{Any short insight why it is happening such?}
% In the three graphs, from left to right, we represent quantitatively the selections on BigHand2.2M, 
% NYU, and ICVL respectively. From the graphs, ..... 
% \textcolor{red}{it seems the UncertainGCN is not performing well. Any Justification?}
% After the ten selection stages, CoreGCN yields the lowest joint errors for all datasets, while UncertainGCN follows or outperforms CoreSet. Therefore, we obtain with CoreGCN 12.3 mm error for ICVL, 27.2 mm for NYU and 28 mm for BigHand2.2M. Under the aforementioned settings, the main differences between our proposed methods and the other baselines can be observed on ICVL where UncertainGCN drops the error below 14 mm at only 300 annotated samples. Summing up, we showed that our proposed AL framework can effectively save the annotation time in the 3D HPE as well.
\subsection{Sub-sampling of Synthetic Data.} 
% In addition to these real image benchmarks for active learning, we also applied our method to collect 
% quality Generative Adversarial Network (GAN) synthetic data set. We choose 
% data generated from Stargan~\cite{choi2018stargan} for face expressions. 
% \textcolor{blue}{For more description about the datasets please refer 
% supplementary material.}

\begin{figure}
    \centering
    \includegraphics[trim=2.5cm 0.2cm 3.5cm 1cm, clip, width=.75\linewidth]{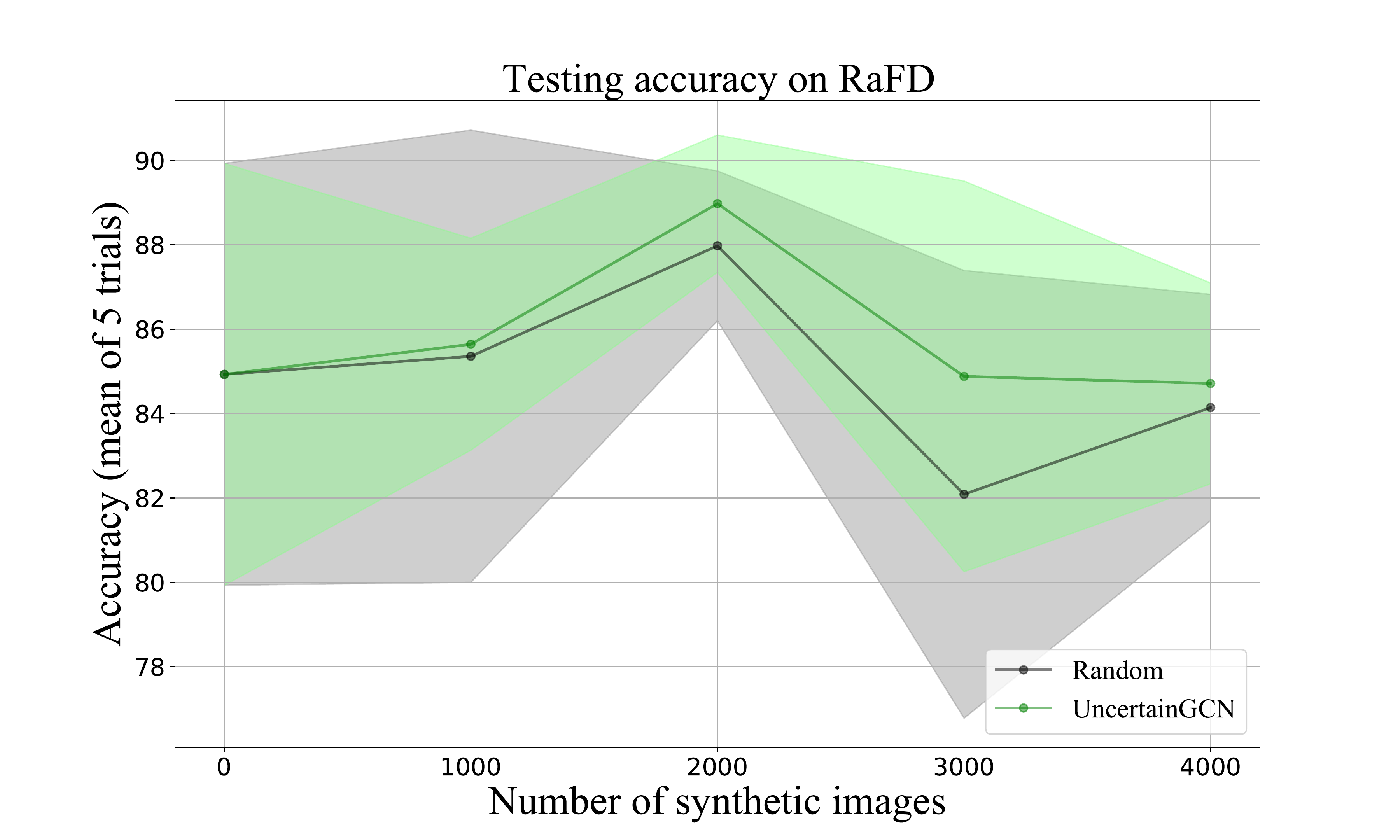}
    \caption{Performance comparison on sub-sampling synthetic data to augment real data for expression classification}
    \label{fig:rafd}
\end{figure}

Unlike previous experiments of sub-sampling real images, we applied our method to select synthetic examples obtained from StarGAN~\cite{choi2018stargan} trained on RaFD~\cite{langner2010presentation} for translation of face 
expressions.
Although Generative Adversarial Networks~\cite{goodfellow2014generative} are closing the gap between real and synthetic data~\cite{ravuri2019seeing}, still the synthetic images and its associated labels are not 
yet suitable to train a downstream discriminative model. Recent study~\cite{bhattarai2020sampling} recommends sub-sampling the synthetic data before augmenting to the real data. Hence, we apply our algorithm to get a sub-set of the quality and influential synthetic examples. The experimental 
setup is similar to that of the image classification which we described in our previous Section. Learner and Sampler remain the same. 
The only difference will be in the nature of the pool images. Instead of real data, we have StarGAN 
synthetic images.
Figure~\ref{fig:rafd} shows the performance comparison of random selection vs our UncertainGCN method in 5 trials. From the experiment, we can observe our method achieving higher accuracy with less variance than commonly used random sampling. The mean accuracy drops for both the methods from the fourth selection stage. 
Only a small fraction of synthetic examples are useful to train the model~\cite{bhattarai2020sampling}. After the fourth stage, we force sampler to select more examples which may end up choosing noisy data.
% RaFD (Radboud Face Dataset ref() - we deploy same ResNet-18 learner but for input images 128x128x3, task is to classify faces into 8 emotional expressions)- we initialise the labelled pool of data with the training set. We consider an unlabelled pool of 8*7200 synthetic data. Similarly to previous experiments we randomly sub-sample a subset synthetic pool from which we further apply random and UncertainGCN selection. This behaviour has been shown in works like...ref() . The synthetic data was generated with MatchGAN ref() .. This proves our versatile improvement in different tasks
%Varying the parameter of the graph and table of comparison vgg11 against resnet18 - robustness to different architectural choices.

\section{Conclusions}
\label{sec:conclusions}
We have presented a novel methodology of active learning in image classification and regression using Graph Convolutional Network. After systematical and comprehensive experiments, our adapted sampling techniques, UncertainGCN and CoreGCN, produced state-of-the-art results on 6 benchmarks. We have shown through qualitative distributions that our selection functions maximises informativeness within the data space. The design of our sampling mechanism permits integration into other learning tasks. Furthermore, this approach enables further investigation in this direction where optimised selection criteria can be combined GCN sampler.
% \clearpage

\section*{Acknowledgements}
This work is partially supported by Huawei Technologies Co. and by EPSRC Programme Grant FACER2VM (EP/N007743/1).

{\small
\bibliographystyle{ieee_fullname}
\bibliography{egbib}
}
\clearpage
% \documentclass[main.tex]{subfiles}
% \begin{document}
% \title{Sequential Graph Convolutional Network for Active Learning \\ (Supplementary Material) }
% \author{Razvan Caramalau$^1$, Binod Bhattarai$^1$ and Tae-Kyun Kim$^{1,2}$ \\
% $^1$Imperial College London, UK\\
% $^2$KAIST, South Korea\\
% {\tt\small \{ r.caramalau18, b.bhattarai, tk.kim\}@imperial.ac.uk}}
\section{Supplementary Material}
% \maketitle
\appendix
\counterwithin{figure}{section}
% \section{Supplementary Material}
% \subsection{Datasets}
\section{Datasets}
% \textcolor{blue}{We give more insights about the data sets which includes their statistics and also some qualitative examples.}
% \noindent \textbf{CIFAR-10/CIFAR-100:}
% \noindent \textbf{SVHN}
% \noindent \textbf{FashionMNIST}
% \noindent \textbf{RaFD}
% \textcolor{red}{add description of these data sets including their statistics and figuer showing sample images}
% \textcolor{green}{I've added information about the datasets. Which other statistics are you interested in? We discussed to add qualitative of icvl, cifar10 (one of the classification dataset) and rafd which are present in Figure \ref{fig:qual}. Do you want me to add images for CIFAR-100, SVHN and FashionMNIST as well?}
\begin{figure}[hbt!]
    \centering
    \includegraphics[trim=0cm 0cm 0cm 0cm, clip, width=.96\linewidth]{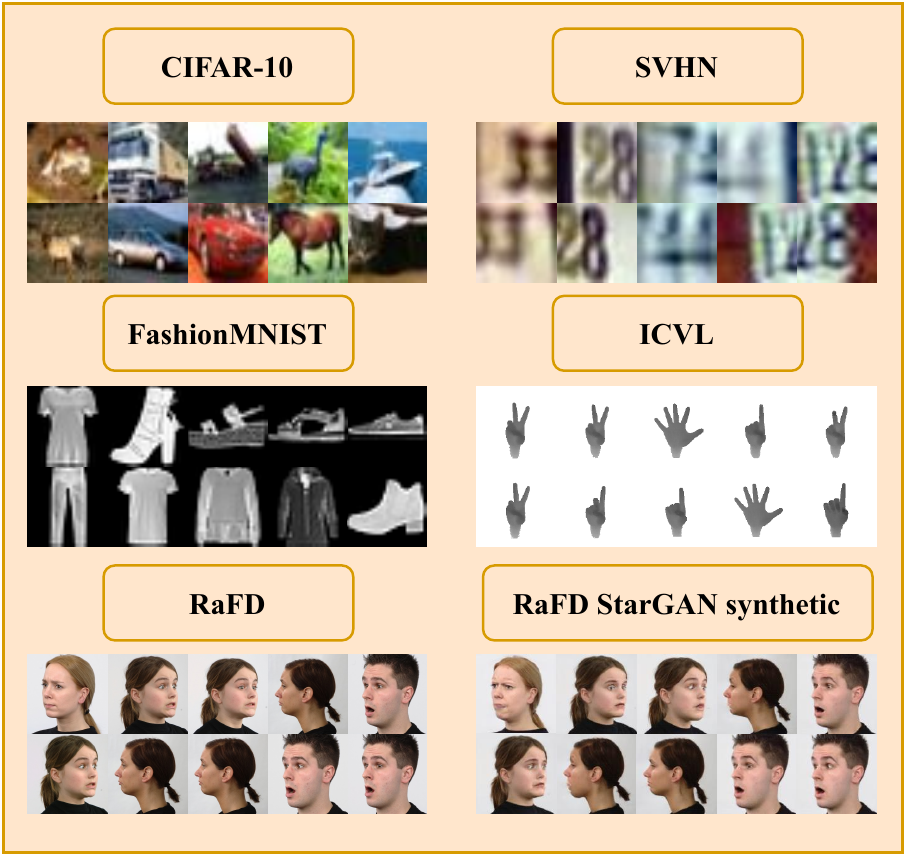}
    \caption{This Figure shows some of the randomly sampled images from the data sets we use to validate our methods. Effectiveness of our method on these diverse characteristics of datasets demonstrate its generic nature.}
    \label{fig:datasets}
\end{figure}
Here, we present a extended description of the datasets we used to evaluate our algorithms and the compared baselines. We evaluated our methods together with the others on four challenging image classification benchmarks: CIFAR-10\cite{cifar}, CIFAR-100\cite{cifar}, FashionMNIST\cite{Xiao2017Fashion-MNIST:Algorithms} and SVHN\cite{Goodfellow2014Multi-digitNetworks}. Each of the datasets has different properties and present new challenges for the active learning framework. FashionMNIST is a grey scale image dataset. Whereas, others 
are RGB image datasets.
\textbf{CIFAR-10} consists of 50,000 images for training and 10,000 for testing. 
% Images are coloured RGB with resolution $32 \times 32$. 
There are 5,000 samples for each of the 10 object categories. 
\textbf{CIFAR-100} is constructed in a similar fashion with the same size of the training and testing set. 
The difference lies in the granularity of the data distribution as 100 classes are categorised 
(500 images corresponding to each class).
% Continuing on RGB images, 
The \textbf{SVHN} dataset represent 10 digit classes with  73,257 train images and 26,032 test images. 
% The resolution is similar to the other two datasets. 
% Finally, the last benchmark, FashionMNIST is composed of greyscale $28 \times 28$ images. 
Finally, \textbf{FashionMNIST} contains training and testing sets of the size 60,000 and 10,000, respectively, with annotations of 10 clothing designs. From an input image resolution perspective, despite FashionMNIST with a 28x28 size, the other datasets have 32x32 scale.

Together with the classification task, we shift the learner's objective to regression. As we tackle the 3D Hand Pose Estimation task, we benchmark our baselines on one of the most challenging, widely been used and first of depth based  datasets, \textbf{ICVL}\cite{icvl}. This is composed of 16,004 images for training and 1,600 for testing. The dataset has a single frontal viewpoint and a wide range of articulation and hand positions. The initial resolution is 320x240, but we pre-process by hand centring and scaling to 128x128.

The last benchmark we deployed in the experiment section is the face expression dataset, Radboud Faces Database
(\textbf{RaFD})\cite{langner2010presentation}. This is formed of 7,200 training images, 800 for each of the 8 expressions.
However, the test set contains only 840 images. Although the initial image dimensions are 256x256x3, for 
efficiency, we downscale them by a factor of 2. As we consider the entire training set as labelled in this
experiment, we generate with StarGAN\cite{choi2018stargan} 57,600 images for the unlabelled set. Similar to the
CIFAR-10 evaluation settings, we initially create a randomly distributed subset $\mathbf{D}_S$ of 10,000 images 
from which we further apply the selection given a budget $b$ of 1,000.

% \subsection{Supplementary: Experiments}
\section{Experiments}
% \appendix

% \beginsupplement
% \newcounter{Figure}[figure]

% \counterwithin{figure}{section} %{figure}{section}

\paragraph{CIFAR-10 imbalanced dataset} In the experimental part, we evaluated quantitatively in a systematic manner the active learning methods over four image classification datasets. Although, before selection, we randomise the unlabelled samples to a subset, the dataset is still relatively balanced to each class distribution. However, this is not commonly the case where there is no prior information related to the data space. Therefore, we are simulating an imbalanced CIFAR-10 in a quantitative experiment.
Beforehand we considered the 50,000 training set as unlabeled, given 5,000 samples for each of the 10 categories. We custom the dataset so that 5 of the 10 classes contain 10 \% of their original data (500 samples each). Therefore, the new unlabelled pool is composed of 27,500 images. The experiment architecture and settings are similar to the one on the full scale.

\begin{figure}[hbt!]
    \centering
    \includegraphics[trim=0cm 0cm 0cm 0cm, clip, width=.96\linewidth]{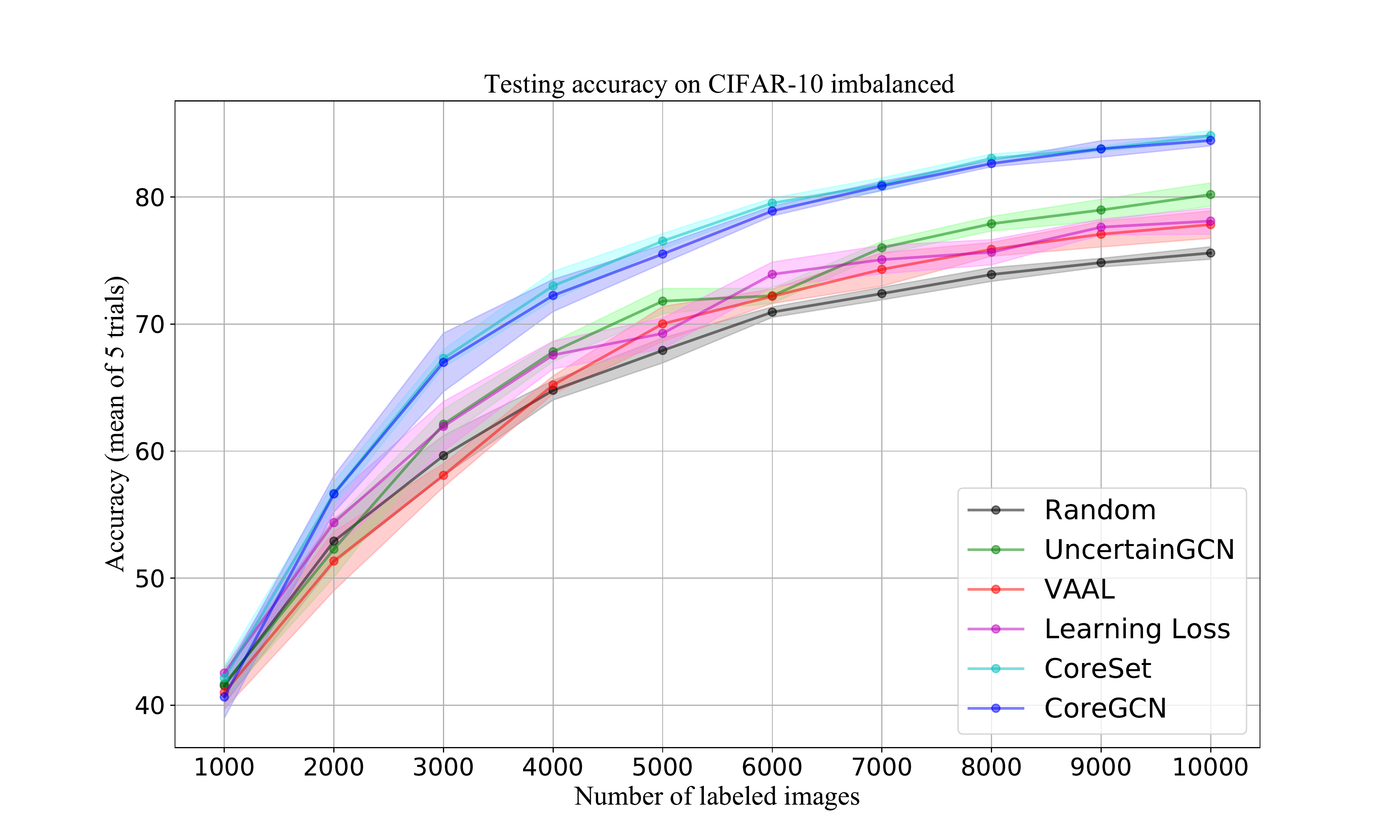}
    \caption{Quantitative results - CIFAR-10 imbalanced dataset}
    \label{fig:cifar10im}
\end{figure}

Figure \ref{fig:cifar10im} shows the progressions of the presented baselines. Our proposed methods, UncertainGCN and CoreGCN, out-stand once again the other model-based selections like VAAL and Learning Loss. UncertainGCN scores 2\% more than those methods with 80.05\% mean average accuracy at 10,000 labelled samples. Meanwhile, CoreGCN achieves 84.5\% top performance together with CoreSet. Thus, the geometric information is more useful in scenarios where the dataset is imbalanced.

% Add the figures in larger size!

\begin{figure}[hbt!]
    \centering
    \includegraphics[trim=0cm 0cm 0cm 0cm, clip, width=.96\linewidth]{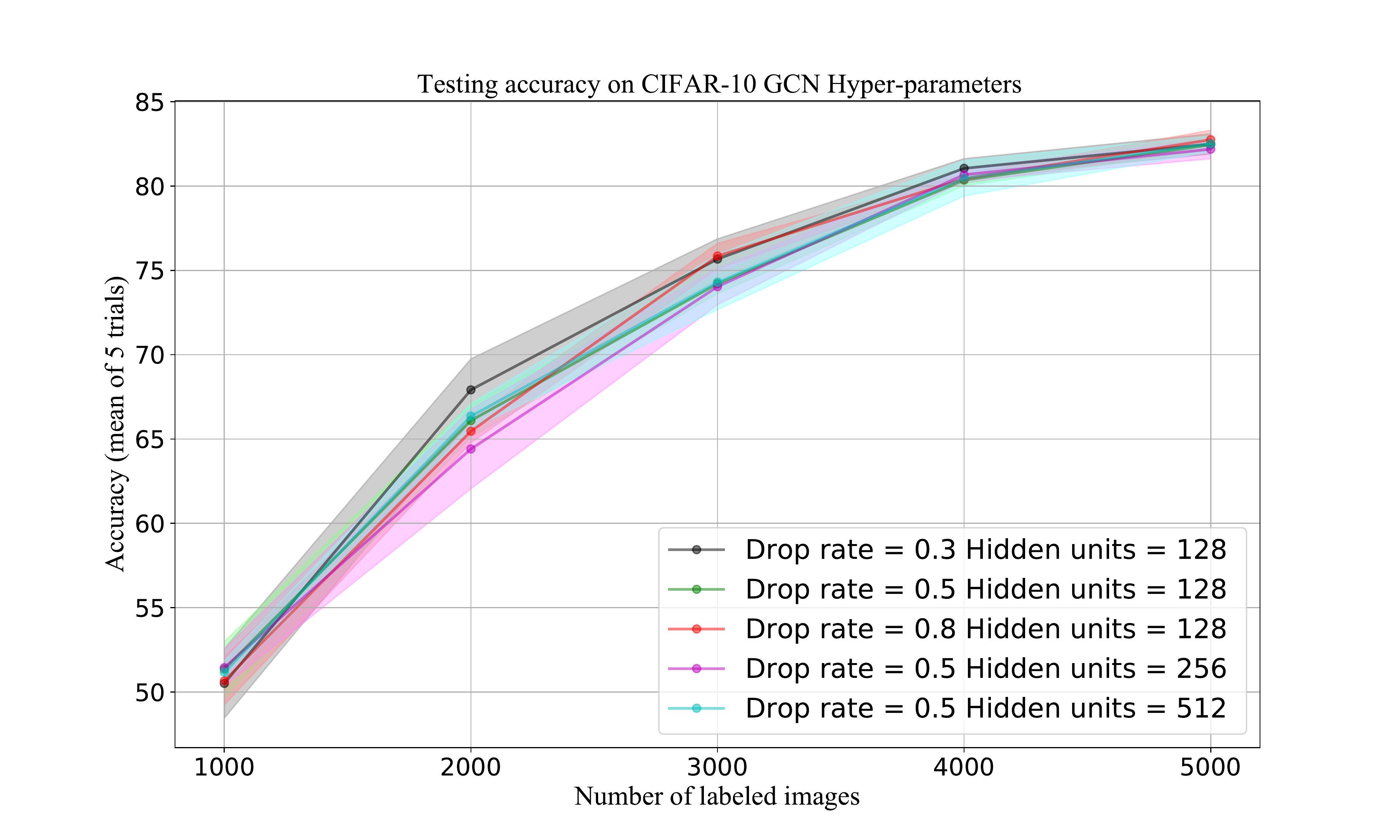}
    \caption{Ablation studies - CIFAR-10 GCN Hyper-parameters tuning}
    \label{fig:hp}
\end{figure}
\paragraph{Ablation study - GCN parameter search} While varying the architectural parameters of the GCN binary classifier, we encountered a poorer
selection with the increase of the Dropout rate from 0.3 to 0.5 or 0.8. However, when changing the
size of the hidden units to 256 and 512, the UncertainGCN sampling was not affected on CIFAR-10. 
This might require further optimisation for different datasets although robustness is being shown.

\begin{figure}[hbt!]
    \centering
    \includegraphics[trim=0cm 0cm 0cm 0cm, clip, width=.96\linewidth]{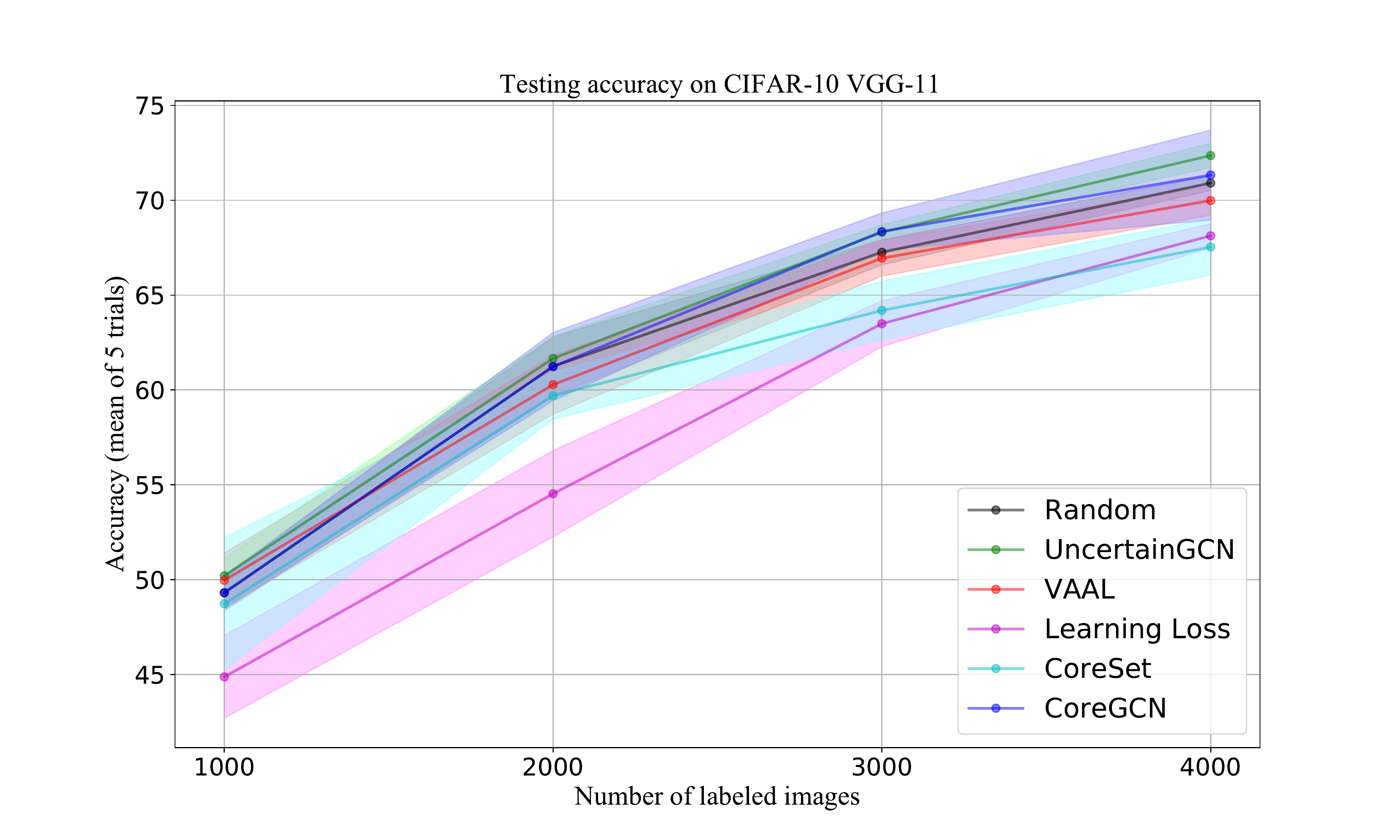}
    \caption{CIFAR-10 Learner VGG-11 - 3 selection stages}
    \label{fig:vgg}

\end{figure}

\paragraph{VGG-11 learner for CIFAR-10 image classification for 3 selection stages} In Figure \ref{fig:vgg}, we modified the architecture of the learner from CIFAR-10 experiment to VGG-11\cite{Simonyan2015VeryRecognition}. Therefore, we analyse how the AL methods are affected in terms of accuracy at the fourth sampling stage. In training the VGG-11 network, we kept the same hyper-parameters. We also had to trace the features after the first four Max Pooling layers for the Learning Loss baseline. Our proposed methods present robustness to this change whilst GCN settings were left unchanged. Hence, they surpass all state-of-the-arts at this early stage. This also demonstrates how the batch size and the feature representation play an important role in the performances of the other baselines. The most affected baseline in this context is CoreSet.

\begin{figure}[hbt!]
    \centering
    \includegraphics[trim=2.5cm 0cm 3.5cm 1cm, clip, width=0.49\linewidth]{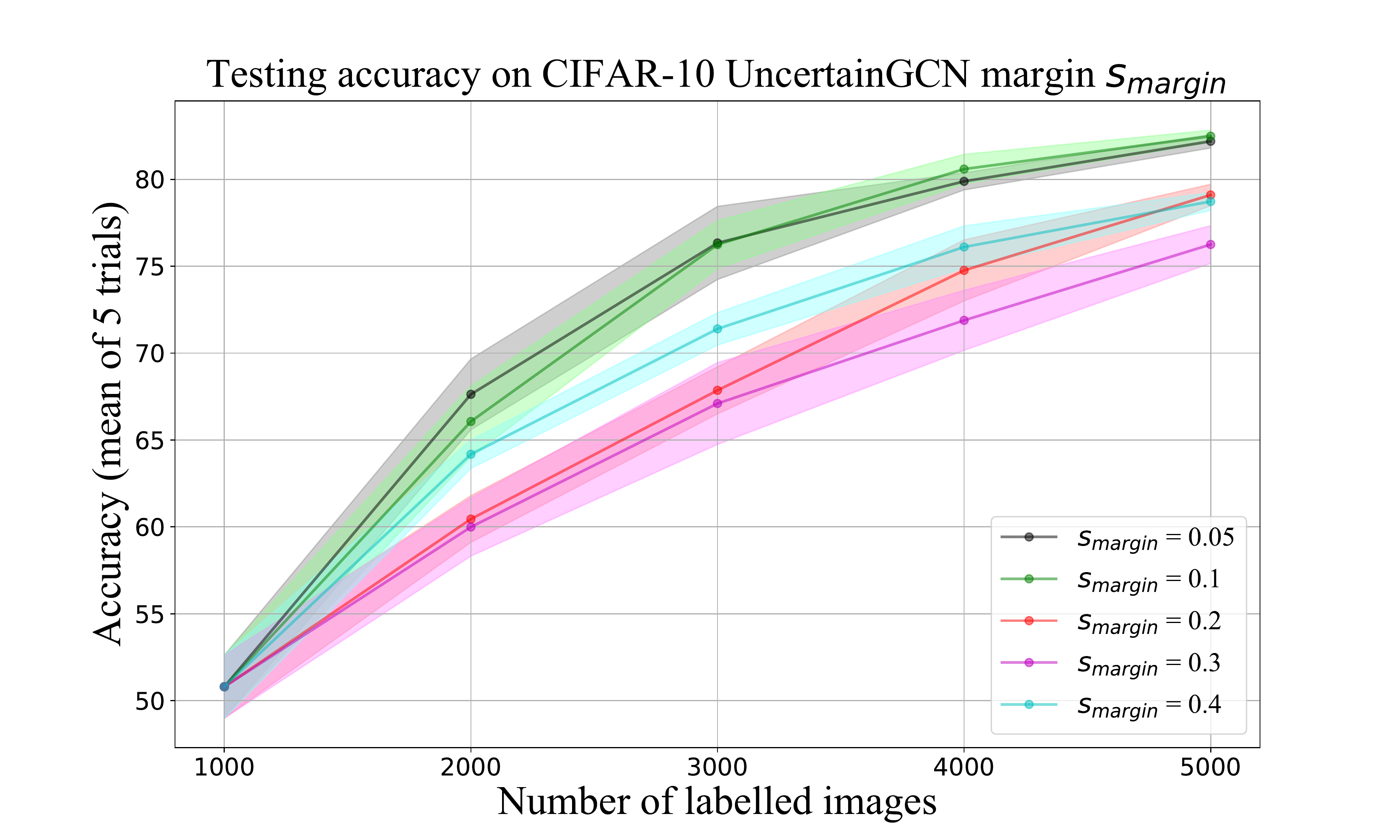}
    \includegraphics[trim=2.5cm 0cm 3.5cm 1cm, clip,  width=0.49\linewidth]{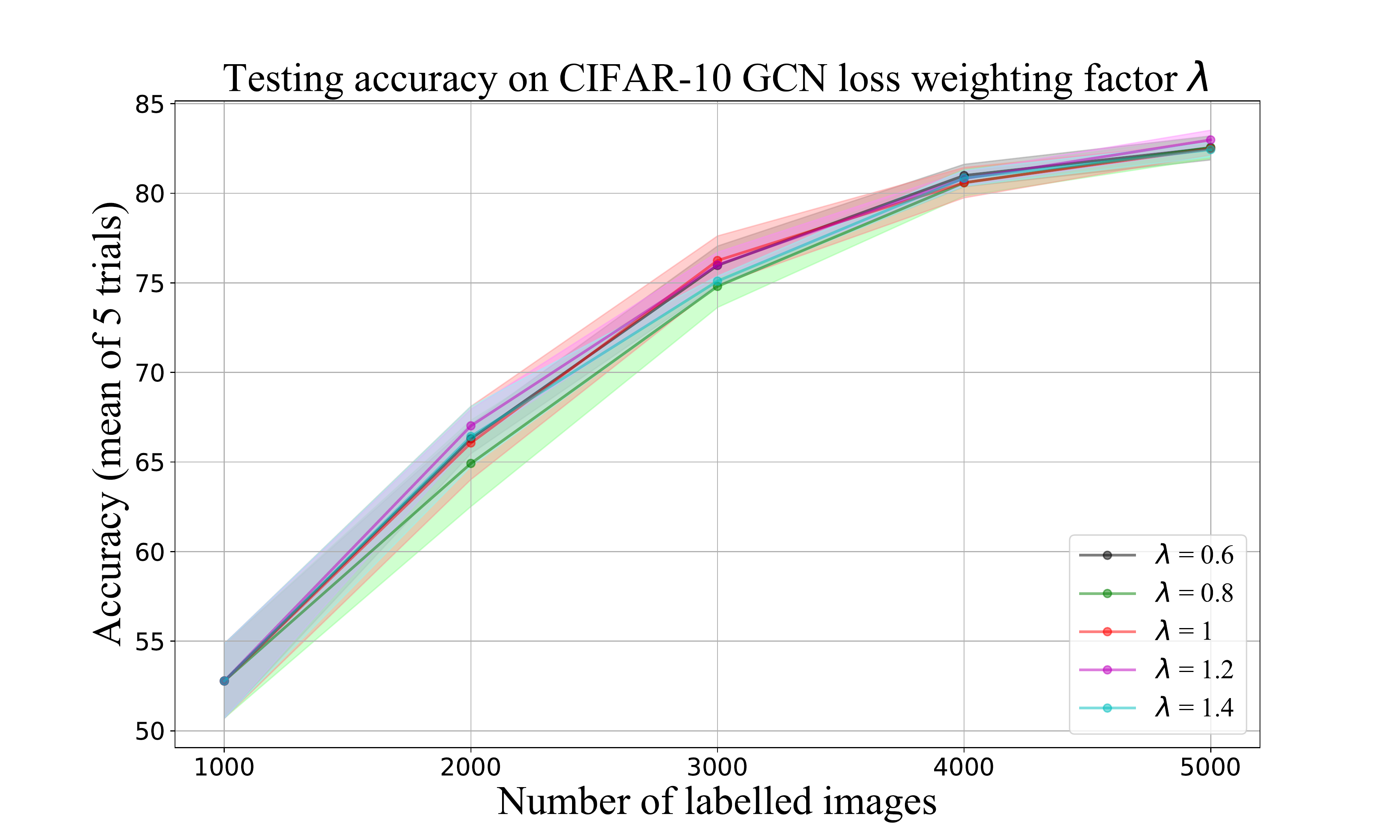}
    \caption{ Hyper-parameter study on UncertainGCN margin ($s_{margin}$) (left) and 
    labelled vs unlabelled data loss weighing factor, $\lambda$ (right) (Zoom in the view)}%
    \label{fig:hyper_param_study}%
\end{figure}

\paragraph{Hyper-parameters Study }
Here, we present the analysis of two important hyper-parameters in the objective of the sampler. 
These are GCN uncertainties margin $s_{margin}$ and  $\lambda$, the labelled vs unlabelled data loss 
weighing factor. Figure~\ref{fig:hyper_param_study} summarises these studies. 
From the Figure, we observe that the performance improves when we decrease margin from 0.4 to
0.1. Afterwards, the performance is stable. This shows that our method is stable in the range of
an optimal margin. Similarly, $\lambda$ influences the performance. However, the drift in
performance is smooth with the change in the value of $\lambda$.

\begin{figure*}
    \centering
    \includegraphics[trim=0cm 0cm 0cm 0cm, clip, width=.9\linewidth]{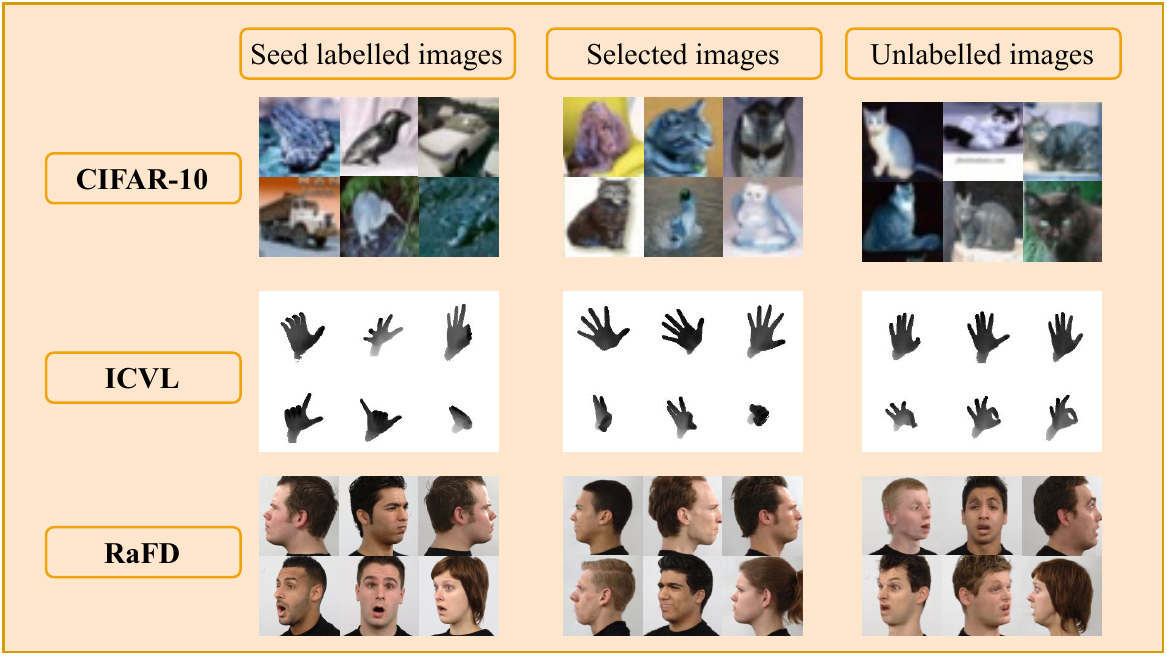}
    \vspace{-0.2cm}
    \caption{Extended qualitative analysis on labelled/unlabelled images at the last selection stage for CIFAR-10, ICVL and RaFD}
    \label{fig:qual}
\end{figure*}

\paragraph{Extended qualitative analysis on the AL method}In Figure \ref{fig:qual}, we extend our qualitative analysis by visualising the initial, the unlabelled and the last selected samples from CIFAR-10, ICVL and RaFD. The last selection stage for CIFAR-10 and ICVL is the 10th, while in the synthetic RaFD experiment is the 4th. The seed labelled images are acquired randomly before the first selection stage. The RaFD seed examples are from the entire training set as the AL selection is applied on StarGAN generated images. 
For all the three benchmarks we evaluated the selected examples with our proposed AL method, UncertainGCN. Although the seed labelled samples for CIFAR-10 are randomly selected, the top query images from the "cat" class consist of difficult examples. On the other hand, the remained unlabelled images present distinguishable features, easy for the learner to predict. These observations have been quantified in the main paper as well. However, in the ICVL dataset case, the selected samples show closer and easier hand articulations compared to the initial labelled set. This is because of the highly complex set that was used as seed examples. The unlabelled images might have a lack of representativeness in the learner's perception after all the 10 
sampling stages. Finally, in the RaFD synthetic sub-sampling process, we can clearly denote the noisy images that were left unlabelled. These present more artefacts than the selected group.

% {\small
% \bibliographystyle{ieee_fullname}
% \bibliography{egbib}
% }

% \begin{figure}
%     \centering
%     \includegraphics[trim=4.0cm 1cm 2.5cm 2.5cm, clip, width=0.40\textwidth]{images/diff_learner.pdf}
%     \caption{Learner comparison [Resnet-18 vs VGG-11] - 4,000 CIFAR-10 labelled images [mean averaged accuracy on 5 trials]}
%     \label{fig:7}
% \end{figure}

% \sout{While varying the architectural parameters of the GCN binary classifier, we encountered a poorer
% selection with the increase of the Dropout rate from 0.3 to 0.5 or 0.8. However, when changing the
% size of the hidden units to 256 and 512, the UncertainGCN sampling was not affected on CIFAR-10. 
% This might require further optimisation for different datasets although robustness is being shown.
% \textcolor{blue}{do you have exact figure for it? If yes move to supplementary. If not remove it}.}

% \end{document}

\end{document}